\documentclass[journal,transmag]{IEEEtran}

\usepackage{textcomp}

\usepackage[noadjust]{cite}
\usepackage[pdftex]{graphicx}
\graphicspath{{../pdf/}{../jpeg/}}
\DeclareGraphicsExtensions{.pdf,.jpeg,.png}
\usepackage{calrsfs}
\usepackage{float}

\usepackage{amsmath}
\usepackage{multirow}
\newcommand{\hl}[1]{#1}
\usepackage{hyperref}

\usepackage{graphicx} 
\graphicspath{ {./} }

\usepackage[rightcaption]{sidecap}
\usepackage{subcaption}

\usepackage{wrapfig}

\usepackage{textcomp}

\begin{document}

\bstctlcite{IEEEexample:BSTcontrol}
\title{Automatic Social Distance Estimation From Images: Performance Evaluation, Test Benchmark, and Algorithm}

\author{\IEEEauthorblockN{Mert Seker\IEEEauthorrefmark{1},
Anssi Männistö\IEEEauthorrefmark{2},
Alexandros Iosifidis\IEEEauthorrefmark{3} and
Jenni Raitoharju\IEEEauthorrefmark{4}}

\IEEEauthorblockA{\IEEEauthorrefmark{1}Unit of Computing Sciences, Tampere University, Finland (e-mail: mert.seker@tuni.fi)}
\IEEEauthorblockA{\IEEEauthorrefmark{2}Unit of Communication Sciences, Tampere University, Finland
(e-mail: anssi.mannisto@tuni.fi)}
\IEEEauthorblockA{\IEEEauthorrefmark{3}Department of Electrical and Computer Engineering, Aarhus University, Denmark (e-mail: ai@ece.au.dk)}
\IEEEauthorblockA{\IEEEauthorrefmark{4}Programme for Environmental Information, Finnish Environment Institute, Jyväskylä, Finland (e-mail: jenni.raitoharju@syke.fi)}
\thanks{M. Seker, A. Männistö, and J. Raitoharju would like to acknowledge the financial support from Helsingin Sanomat foundation, project "Machine learning based analysis of the photographs of the corona crisis". A. Iosifidis acknowledges funding from the European Union’s Horizon 2020 research and innovation programme under grant agreement No 957337. 
Corresponding author: Mert Seker (e-mail: mert.seker@tuni.fi).}}

\markboth{Seker \it{et al}.: Automatic Social Distance Estimation From Images}{Seker \it{et al}.: Automatic Social Distance Estimation From Images}

\maketitle

\begin{abstract}
The COVID-19 virus has caused a global pandemic since March 2020. The World Health Organization (WHO) has provided guidelines on reducing the spread of the virus and  one of the most important measures is social distancing. Maintaining a minimum of one meter distance from other people is strongly suggested for safety.
While social distancing regulations can slow down the spread of the virus, they also directly affect a basic form of non-verbal communication, and there may be deeper and longer term impacts on human behavior and culture that remain to be analyzed in social sciences and proxemics studies. To obtain quantitative results for such studies, it is necessary to analyze large numbers of personal and/or media photos. Several methods have been proposed for automatic social distance monitoring, but they are not directly applicable for analyzing such general photo collections, where the variations in the imaging setup are large. Furthermore, in such studies the interest shifts from monitoring whether the recommended safety measures are followed to more subtle differences in social distances. Currently, there is no suitable test benchmark for developing such algorithms. Collecting images with measured ground-truth pair-wise distances between all the people using different camera settings is cumbersome. Furthermore, performance evaluation for social distance estimation algorithms is not straightforward and there is no widely accepted evaluation protocol. In this paper, we provide a dataset of varying images with measured pair-wise social distances under different camera positions and focal length values. We suggest a performance evaluation protocol and provide a benchmark to easily evaluate social distance estimation algorithms. We also propose a method for automatic social distance estimation. Our method takes advantage of object detection and human pose estimation. It can be applied on any uncalibrated single image as long as focal length and sensor size information are known. 
The results on our benchmark are encouraging with 92\% human detection rate and only 28.9\% average error in distance estimation among the detected people.  
\end{abstract}

\begin{IEEEkeywords}
Social Distance Estimation, Person Detection, Human Pose Estimation, Performance Evaluation, Test Benchmark, Proxemics
\end{IEEEkeywords}

\maketitle

\section{Introduction}
\label{sec:intro}

\hl{Social distances are a part of non-verbal human communications and, naturally, there are personal and cultural differences in how people feel about their personal space and interpret the interpersonal distance in different situations. The research field under social studies concerning these phenomena related to space is known as \emph{proxemics} \mbox{\cite{hall1968proxemics}}. Despite the long history of studies in the field \mbox{\cite{hiddendimensions, cook1970proxemics, harrigan2005proxemics}}, it remains difficult to carry out quantitative analysis on the actual social distances in the natural situations outside of monitored test conditions, e.g., when people are spending their free time with their families. One way to approach this problem is \mbox{\emph{visual social distancing}} (VSD), where the interpersonal distances are automatically measured from the images or videos. A comprehensive overview of the VSD problem, including the main challenges and connections to social studies, is provided in \mbox{\cite{vsd}}.}

\hl{Social distancing has recently received a lot of attention due to t}he outbreak of SARS-CoV-2 virus \cite{covid19} \hl{that} was declared as a global pandemic by the World Health Organization (WHO) in March 2020.  The pandemic, also known as the COVID-19 pandemic is still ongoing as of \hl{May 2021} and there has been a total of \hl{about 164} million confirmed cases and \hl{3.4} million deaths worldwide within the period of December 2019-\hl{May} 2021 \cite{coviddata}. Social distancing plays an important \hl{role} in slowing down the spread of the virus. \hl{WHO} recommends to stay at least one meter apart from other people in order to reduce the risk of infection \cite{who}. Automatically monitoring the social distances is important for safety reasons, but it is also interesting as a phenomenon that has globally changed basic human behavior \cite{zhang2021memory, dicorrado2020physicalactivity, eden2020media}. After the pandemic eases, there are many interesting research questions \hl{in proxemics and other fields} to look into: how the social distancing affected every-day life, what kind of significant differences were there between different countries, can the differences be linked to the spreading speed, will there be any long-term changes that will stay after the pandemic. 

\begin{figure}[t]
    \centering
    \includegraphics[width=1\columnwidth]{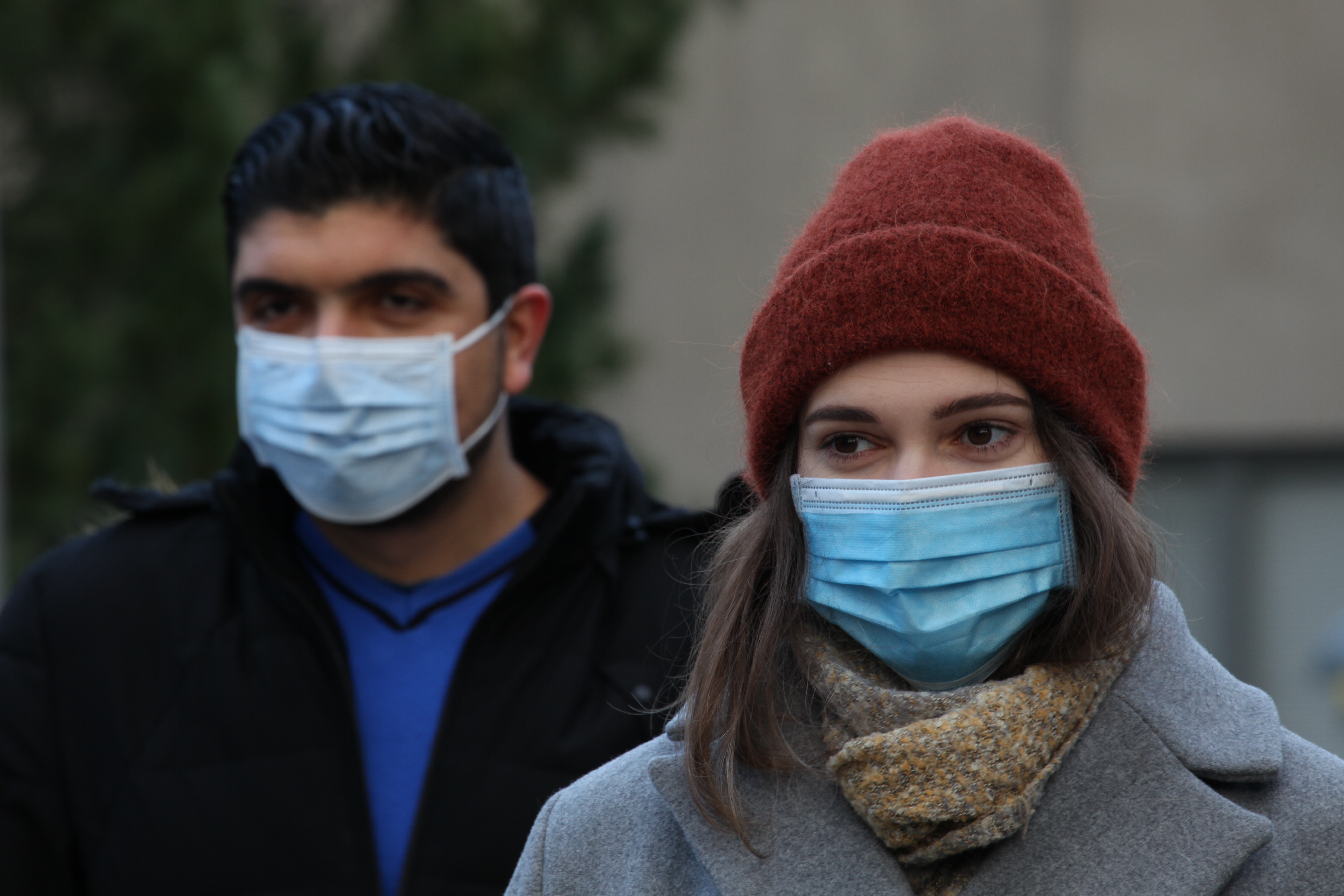}
    \caption{\hl{An example of an image that represents a style, which is common in personal and media photography, but not in monitoring.}}
    \label{fig:mediaexample}
\end{figure}

\hl{While there are methods and sensors available for automatic monitoring and surveillance of social distances \mbox{\cite{nguyen_saputra_van}}, the analysis of deeper and longer term social and cultural impacts of the social distancing regulations requires looking into different source data, such as people's personal photo collections and pictures published in newspapers and magazines. For monitoring purposes, it is possible to use fixed camera setup and location, take videos or simultaneous images from multiple viewpoints, and use additional sensors such as depth or thermal cameras. All these can make the social distance estimates more accurate but are not available for typical personal and media photos that are not taken with a fixed setup, but have varying parameters such as focal length, sensor size, lighting conditions, and pitch angle. An example of an image that could be found in a personal or media photo collection, but not in a monitoring or surveillance setup is shown in Fig.~\mbox{\ref{fig:mediaexample}}.
At the same time, in such social and proxemics studies the focus shifts from monitoring whether people are obeying the regulations to more subtle differences in the social distances and how they are represented in the media.}

\hl{During the pandemic, most effort has been understandably on the monitoring side, and} currently there is no suitable benchmark for developing and testing algorithms for \hl{accurate social distance analysis from single images having varying camera parameters.} This can be due to the laboriousness of gathering varying images with measured pair-wise distances between humans. At the same time, there is no clear protocol for measuring the algorithm performance \hl{in this task}. To address these lacks, we provide a social distance evaluation test benchmark including a protocol for mapping the detected pair-wise distances into the corresponding ground truth distances, a suggested overall performance metric, and 96 test images taken with varying setups: indoors-outdoors, sitting-standing, varying camera angles using 2 different cameras and 7 different focal lengths. \hl{The photos were taken by a professional photojournalist to follow the typical media photography style.}  
We publish \hl{also} easy-to-use codes for evaluating novel methods and  make it easy to integrate additional test photos.

We also propose a social distance estimation algorithm \hl{that can be applied on any uncalibrated single image taken by a regular camera as long as focal length and sensor size are known.}  It combines object detection and human pose estimation with projective geometry \hl{using} image parameters (focal length, sensor size) and pixel locations. While the results are promising, we also point out some of the main remaining challenges for future development.

The rest of the paper is organized as follows. Section~\ref{sec:related} introduces related work on social distancing and automatic distance evaluation. Section~\ref{sec:testdataset} describes the provided test benchmark and the proposed evaluation protocol. Our method for automatic social distance estimation is described in Section~\ref{sec:method}. Section~\ref{sec:experimental} provides our experimental setup and results and, finally, 
Section~\ref{sec:conclusion} concludes the paper.

\section{Related Work}
\label{sec:related}

Effectiveness of social distancing on slowing down the spread of the COVID-19 virus have been widely studied \cite{voko,sun_zhai_2020,prem_liu_russell_kucharski_eggo_davies_group_jit_klepac_2020,courtemanche_garuccio_le_pinkston_yelowitz_2020,abouk_heydari_2021,balasa_2020} and these studies have shown that social distancing measures are successful in reducing the growth rate of the virus. Therefore, monitoring and regulating the social distancing behaviour between people plays a crucial part in dampening the effects of the virus. In addition to directly effecting the virus spread, social distancing has globally changed human behavior and interactions leading to different side-impacts, e.g., on mental health \cite{ford2020psychological, jacob2020crosssectional}, physical activity \cite{dicorrado2020physicalactivity, jacob2020crosssectional}, mood and memory \cite{zhang2021memory}, and media consumption \cite{eden2020media}. Such impacts and their cross-cultural \cite{alhasan2020crosscontinental, alhasan2020comparative, doogan2020twitter} and cross-sectional \cite{jacob2020crosssectional, lee2021southkorea} differences continue to draw attention from researchers in many fields.  

\begin{figure*}[t]
    \centering
    \includegraphics[width=1\textwidth]{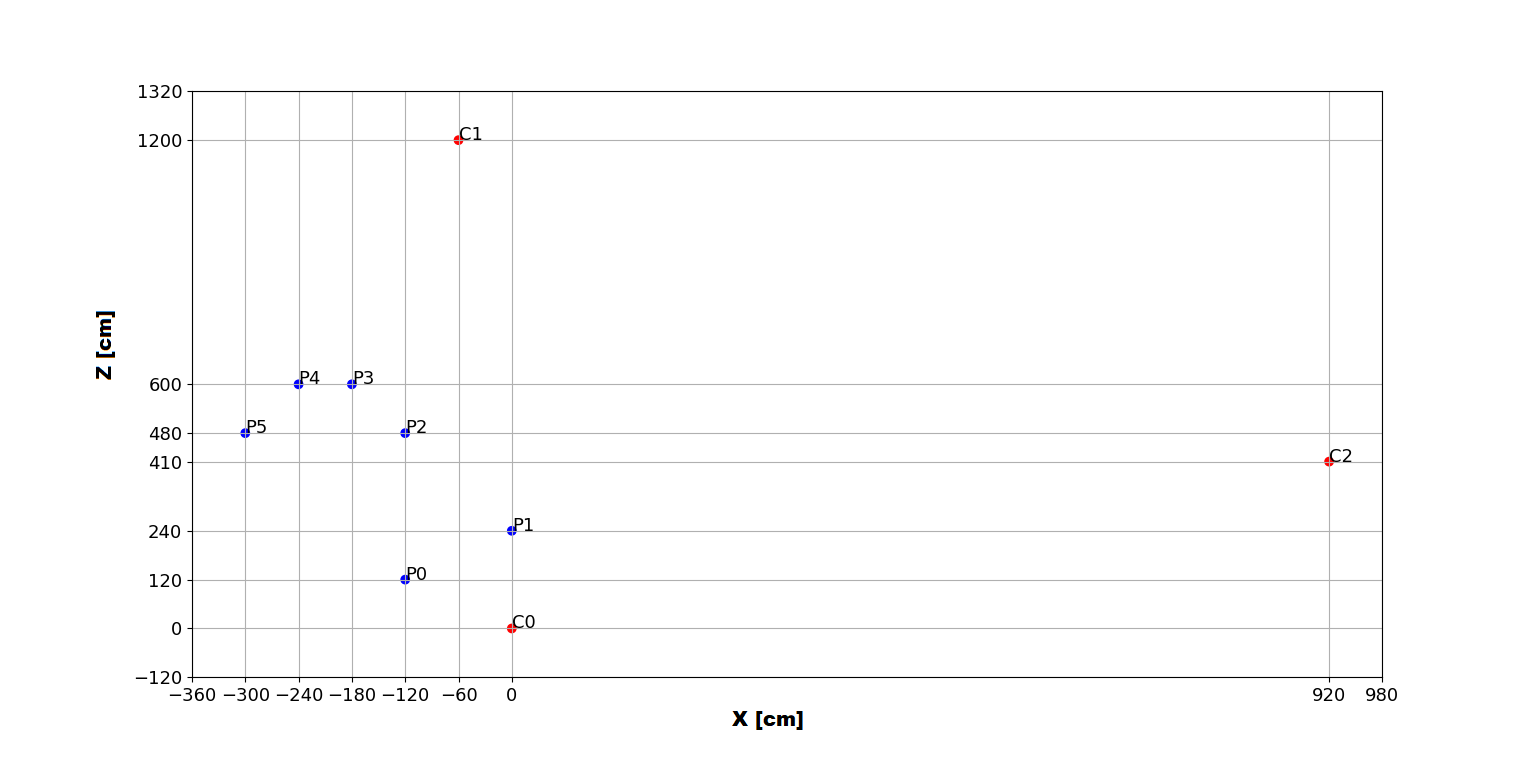}
    \caption{Birdseye view of the outdoor photo shoot. \hl{The ground truth locations of the people and cameras are given in blue and red dots, respectively.}}
    \label{fig:birdeye1}
\end{figure*}
\begin{figure*}[t]
    \centering
    \includegraphics[width=1\textwidth]{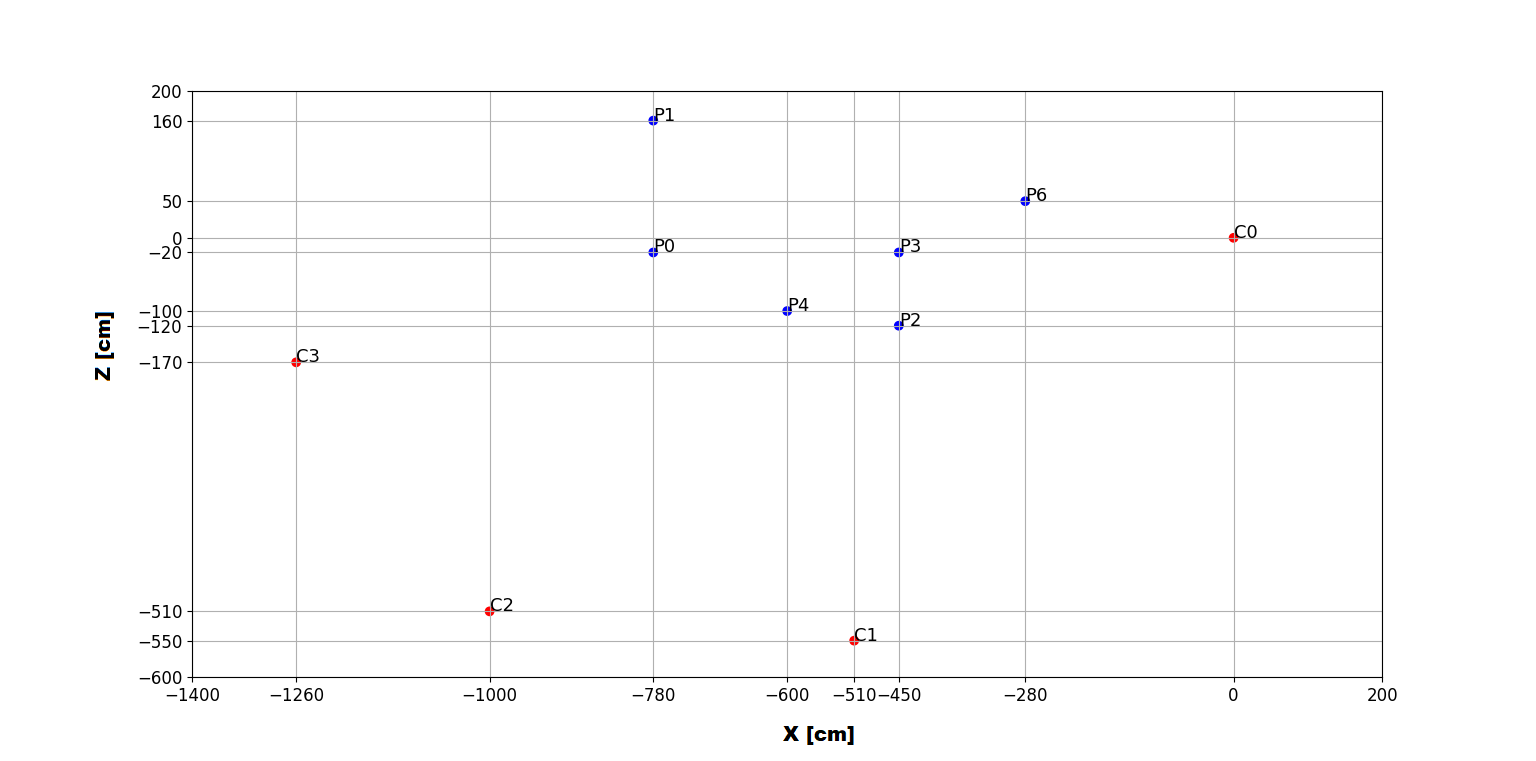}
    \caption{Birdseye view of the indoor photo shoot. \hl{The ground truth locations of the people and cameras are given in blue and red dots, respectively}}
    \label{fig:birdeye2}
\end{figure*}

Social distance monitoring for safety reasons can be eased by automatic social distance estimation from images and videos.
Recent advancements in machine learning, computer vision, thermal and ultrasound technologies have made this task possible. A comprehensive survey in \cite{nguyen_saputra_van} explores the wide array of current technologies that can be used to monitor and encourage social distancing.  A mobile robot that incorporates a 2D lidar to autonomously navigate in crowded environments without colliding with people was described in \cite{covidrobot}. The robot uses a RGB-D camera to detect people and it estimates the distance between the detected people by using the visual and depth information. A commercial pedestrian tracking system was used in \cite{pedestriantracker} to detect passengers in crowded environments and estimate the distances between them by using a graph based approach. The method was employed in a Dutch train station and the results show that the average number of pedestrians and social distance violations in the station have decreased after the pandemic. 

A study in \cite{ahmed_ahmad_rodrigues_jeon_din_2021} proposed using a deep learning based model with YOLOv3 \cite{yolov3} as its backbone to monitor social distancing violations from overhead view cameras.
In \cite{yolodeepsorttracker}, the authors used YOLOv3 \cite{yolov3} and DeepSort \cite{deepsort1,deepsort2} to detect bounding boxes of people in RGB images and by utilizing these boundary boxes, they detected the cases of social distance violations. Another study \cite{drone} used an autonomous drone and a YOLOv3 model \cite{yolov3} trained with a custom dataset. The model processes images from the live feed of the drone's RGB camera. The drone detects whether or not a person is wearing a face mask and monitors social distance violations between the people who are not wearing masks. 

A work in \cite{proxemics} proposed to use skeleton keypoints generated from human body pose estimation algorithms \cite{openpose1,openpose2,openpose3,openpose4} to estimate the distance between people from uncalibrated images. The authors used manual tuning to estimate the homography matrix \cite{proj_geo} of an image plane and then used leg, arm, and torso lengths of the people alongside with the homography matrix to draw a safe space circle underneath every detected person. Then, any collision between the estimated safe space circles was reported as a social distance violation. \hl{Similarly, the work in \mbox{\cite{interhomines}} also takes advantage of manual homography matrix calibration to estimate social distances for fixed cameras. Separating the work from \mbox{\cite{proxemics}}, bounding boxes obtained from the object detection model \mbox{\cite{centernet}} and the height of these boxes were used as reference points to estimate the locations of the people. Moreover, a small CNN is used to detect the feet locations even when they are not visible. The output of this CNN is used to correct the height of the bounding boxes in cases of occlusions.
Another similar study in \mbox{\cite{yang2020visionbased}} also used bounding boxes obtained from object detectors \mbox{\cite{yolov4, fasterrcnn}} to estimate locations of the people from surveillance camera footage by using the homography matrix that is calculated from the known extrinsics. The work in \mbox{\cite{monoloco}} used a feed forward neural network that was trained on the intrinsic parameters of the camera and the keypoints obtained from a pose estimation model. The model outputs the predicted 3D locations as well as the orientations of the detected people. While detecting safe distance violations, not only the proximity but also the orientation of the people with respect to one another is considered.}  \par

\hl{Most of the introduced works approach automatic social distance estimation as a monitoring or surveillance task, where the goal is to prevent social distance regulation violations. To this end, they apply additional sensors, use predefined camera settings, and/or manually define a homography matrix for a certain environment. While such approaches can improve the social distance estimation accuracy, they are not feasible when the purpose is to analyze the impacts of social distances in personal or media photo collections.} 

Moreover, the above-mentioned studies approach the automatic social distance estimation problem as a binary classification problem where they aim to classify the pair-wise distances between people either as safe or unsafe, depending on a given threshold. Classifying distances in a binary manner has a high tolerance for distance estimation errors. For example, if the threshold for safe distance is set to 2 meters, the actual distance between a pair of people is 1.9 meters, and a method estimates that distance as 0.1 meter, the percentual distance estimation error would be 94.7\%, but a binary classification approach would still correctly label the situation as a social distance violation. \hl{Furthermore}, the binary approach does not provide any additional information on the severity of the violations in different situations which may be relevant information for subsequent analysis. 

A common pattern observed in most of the machine learning based social distance estimation methods (with the exception of at least \cite{proxemics, monoloco} that use keypoints of the human body) is that they rely on the bounding boxes drawn by object detectors to detect social distance violations. Although the current object detectors are accurate in detecting objects, the bounding boxes are generally loosely drawn around these objects. Thus, it is not reliable to use only the bounding box information for estimating exact distances between people as it is not possible to infer accurate 3D location estimates from the bounding boxes alone. Therefore, we aim to estimate exact 3D locations of all the people in uncalibrated RGB images with respect to the camera by using the information extracted from the human body skeleton detected by body estimation algorithms. Moreover, we also incorporate an object detection model for people detection. However, the purpose of the people detection in our approach is to only detect the false positives in skeleton keypoints, when they are drawn on non-human objects. \par

The method in \cite{proxemics} is the most similar to our method as is also uses body poses. In \cite{proxemics}, manual input is used to estimate the homography matrix of the image plane to the ground plane. The method is evaluated on surveillance camera footage and the task is approached as a binary classification problem. It is feasible to manually set the homography matrix of surveillance cameras as these cameras are generally non-moving and stable. Contrary to this, we want our method to be fully automatic as we aim to estimate distances in images taken in different locations with different cameras. \hl{Instead of requiring manual input to estimate the homography as} the study in \cite{proxemics}, we assume \hl{that we can find keypoint pairs that are parallel to camera's sensor plane} and we use the image parameters, i.e., focal length and sensor size in our distance estimation.

\hl{For the developing and testing social distance estimation methods, it is important to have image datasets that have a suitable setup and ground-truth for the task. The previous works have used datasets such as Epfl-Mpv-VSD \mbox{\cite{epflmpv}}, Epfl-Wildtrack-VSD \mbox{\cite{epflwildtrack}} and OxTown-VSD \mbox{\cite{oxtown}}. These dataset include videos taken by surveillance cameras with fixed extrinsic and intrinsics and they do not include manually measured ground truth locations and distances. Instead, the locations of the people are estimated by making use of the annotation boxes that were drawn on the people. The pixel locations of these annotation boxes are used as a reference point to estimate the subjects\mbox{’} locations by taking the extrinsic parameters into account. This means that these locations are not exactly ground truth, but estimations based on the known extrinsics and the pixel locations of the manually annotated person bounding boxes. Furthermore, since exact body parts are not annotated and the annotations are only in bounding box format, it is not feasible nor possible to accurately match the detected people with the given ground truth people when there are multiple overlapping boxes.
Moreover, only the people that are passing on a certain region of interest are annotated. 

Due to the aforementioned reasons, the existing datasets are not suitable for evaluating methods that aim at estimating distances in general photo collections and are not manually tuned for a specific camera and environments. Furthermore, the approximate person annotations and location estimates do not allow accurately measuring the distance estimation performance, but are only suitable for detecting coarse violations in social distancing recommendations. While this may be sufficient for surveillance purposes in fixed environments, more accurate ground-truth and annotations are needed for evaluating methods aiming at detecting subtle changes in long-term social distancing behavior in varying environments. In the following section, we introduce our novel dataset that addresses the mentioned drawbacks of the existing datasets.
}

\section{KORTE SOCIAL DISTANCE ESTIMATION BENCHMARK}
\label{sec:testdataset}

We provide a test benchmark for facilitating research in automatic social distance evaluation. We propose a performance evaluation protocol and provide 96 test images with ground-truth pair-wise distances. While the number of images is too low for training fully learning-based systems, it provides a varied test setup. All the evaluation codes along with the test photos are publicly available at \href{https://doi.org/10.23729/b2ea87e6-b845-46b8-abf3-cdbe299ce8b0}{https://doi.org/10.23729/b2ea87e6-b845-46b8-abf3-cdbe299ce8b0}. It is also easy to complement the benchmark with additional images by following the proposed annotation format and using the provided evaluation protocol.

\subsection{Test Photo Collection}

We collected test photos in two separate photo shoots. The first photo shoot was organized outdoors at Tampere University campus in December 2020. Every person was standing. The second photo shoot was organized indoors at Tampere University campus in January 2021 with people sitting around tables. In both photo shoots we had five volunteer test subjects. We followed the COVID-19 restrictions at the time: everyone was wearing a mask and we were less than 10 people gathering. As an additional safety measure, we placed to closest distances from each other only people who meet regularly anyway because they share working space or live together. Every test subject signed an agreement allowing to use their images for research purposes. Any bypassers in the images were censored out to respect their privacy and because their exact positions were unknown. \hl{The photos were taken by a professional photojournalist}.

During the photo shoots, test subjects stayed on the same known positions, while the photographer changed his position and used multiple cameras and lenses at each spot. Fig.~\ref{fig:birdeye1} shows the birdseye view of the first outdoor photo shoot. P0, P1, P2, P3, P4, P5 are the locations of the 5 test subjects and C0, C1, C2 are the camera locations. For the first photo shoot, P0, P1, P2, P3, P4, P5, C0 and C1 were all on the same ground plane, while C2 was at a balcony with a height of 230 cm relative to the ground plane that all of the other locations were at. Fig.~\ref{fig:birdeye2} shows the birdseye view of the second photo shoot. P0, P1, P2, P3, P4, P6 are the locations of the 6 people and C0, C1, C2, C3 are the camera locations. The unit of the x and y axis labels is centimeters. For the second photo shoot, all of the locations were on the same ground plane. \hl{The ground truth locations of the cameras and the test subjects were measured and maintained exploiting tiles on the ground/floor that were equal in size. Only the camera location C2 in the outdoor photo shoot was not on the same tiling and, therefore, it was measured separately using a measuring tape.
}

\hl{We do not report the exact pitch angles, and they were not fixed in the photo shoots. Due to the camera positions, pitch angles are close to zero in most of the images except for the 16 photos taken from camera position C2 in the outdoor photo shoot, where the camera was at an elevated position. We believe that our dataset represents a typical media or personal photo collection with respect to the pitch angles, but it should be noted that methods performing well on our dataset (especially if they rely on the zero pitch angle assumption) may not perform equally well on extreme pitch angles such as overhead images.}

\begin{figure*}[h]
    \centering
    \includegraphics[width=0.32\linewidth]{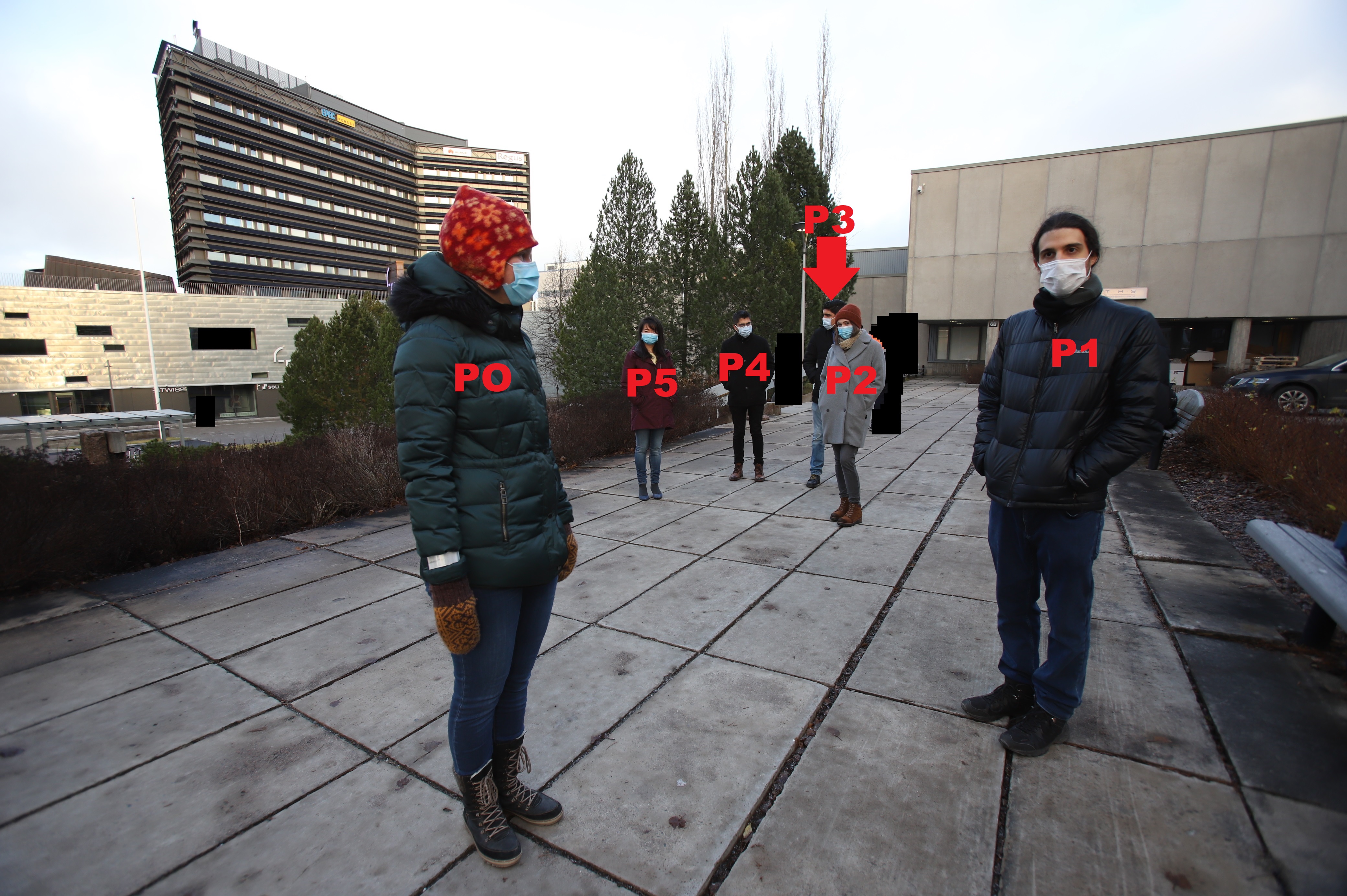}
    \includegraphics[width=0.32\linewidth]{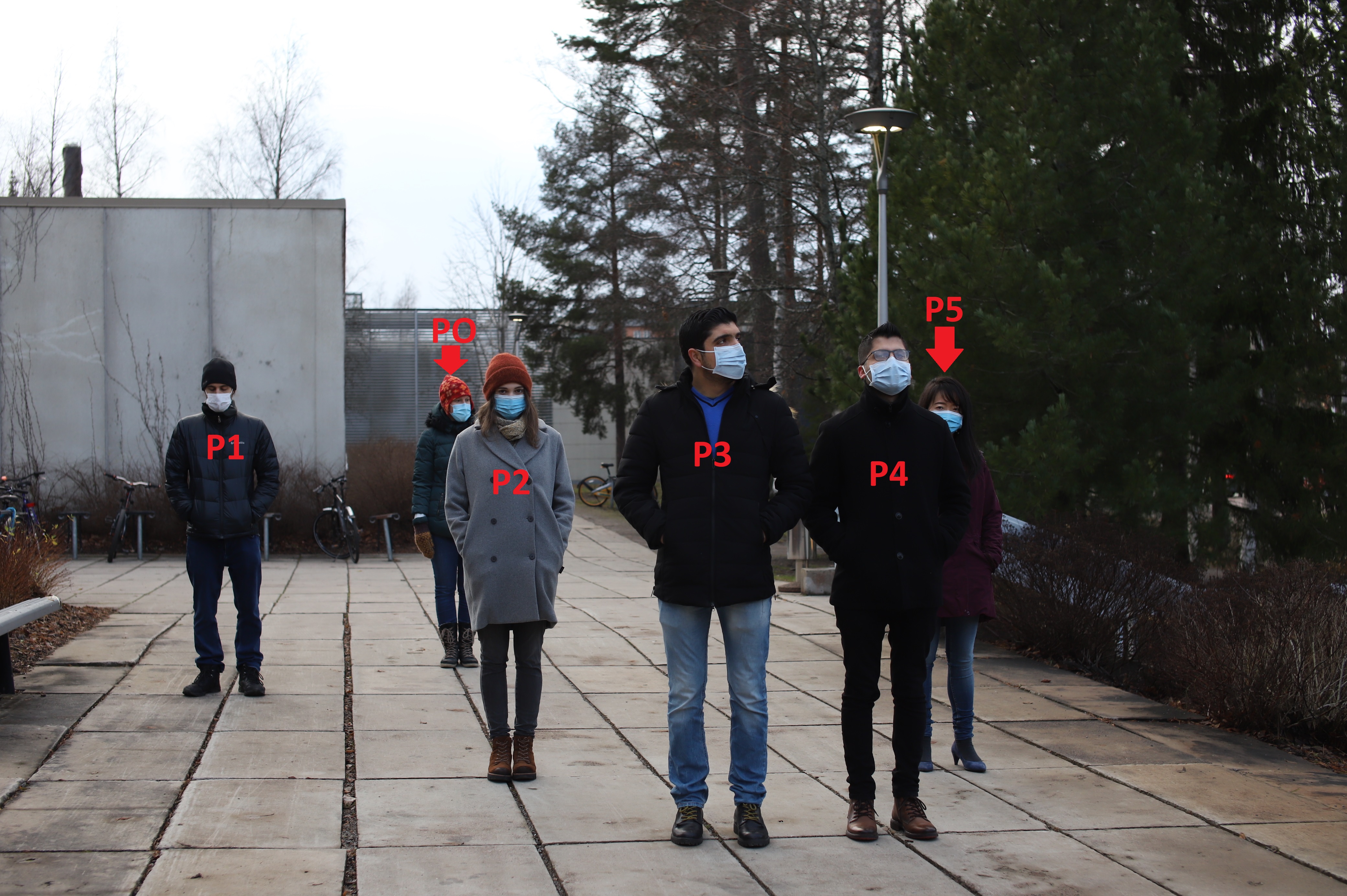}
    \includegraphics[width=0.32\linewidth]{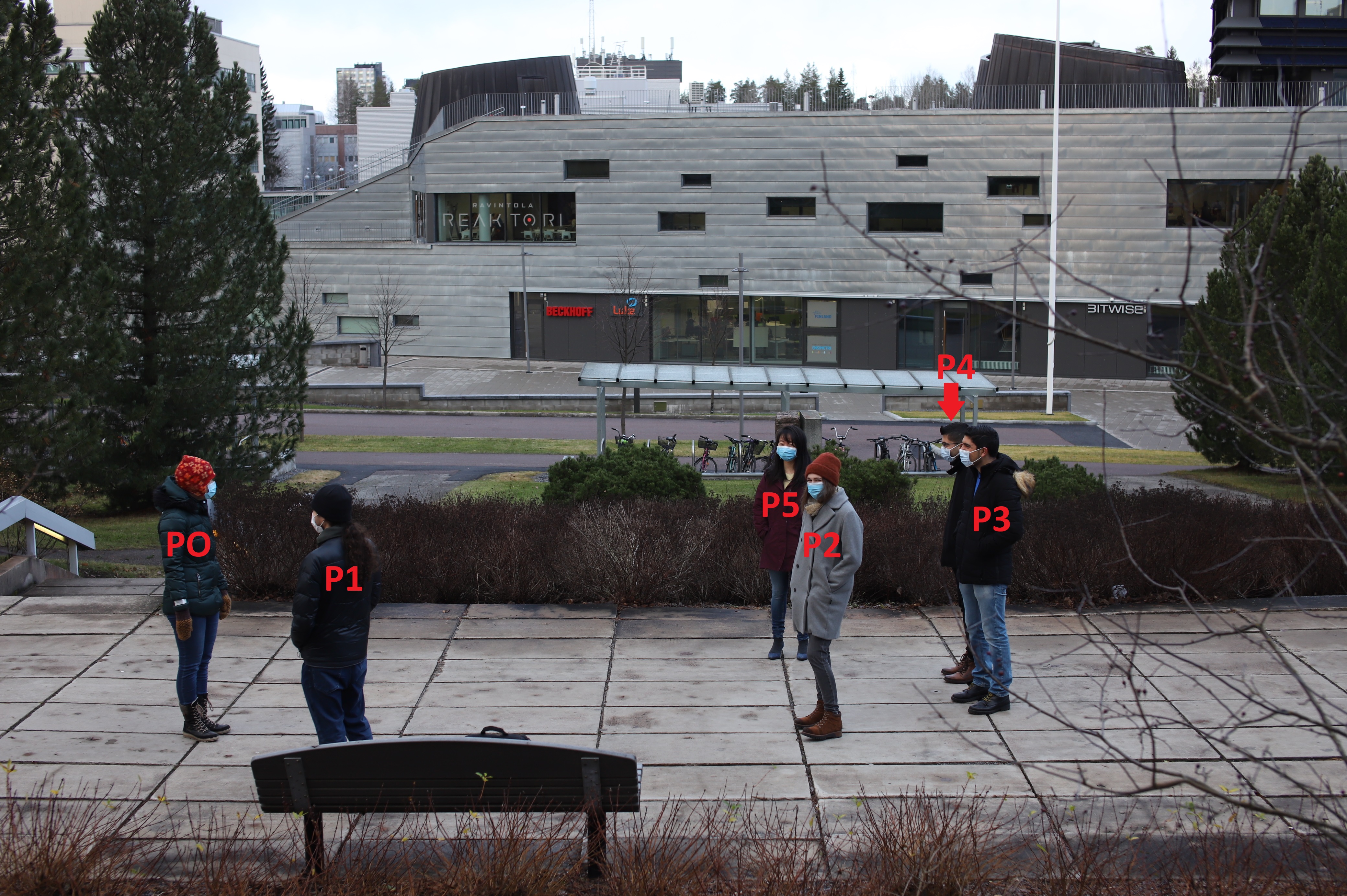} \\
    \vspace{5pt}
    \includegraphics[width=0.24\linewidth]{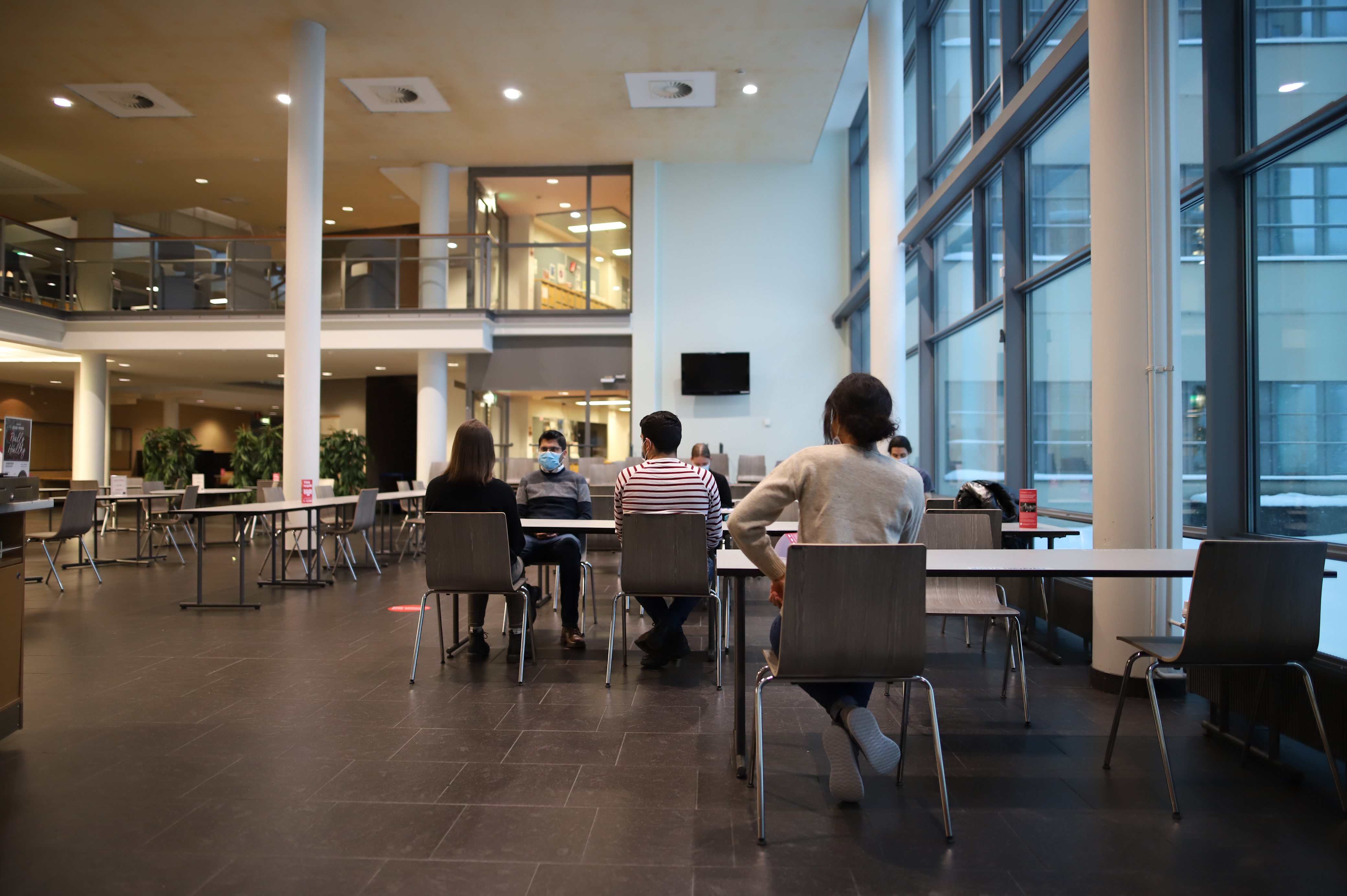}
    \includegraphics[width=0.24\linewidth]{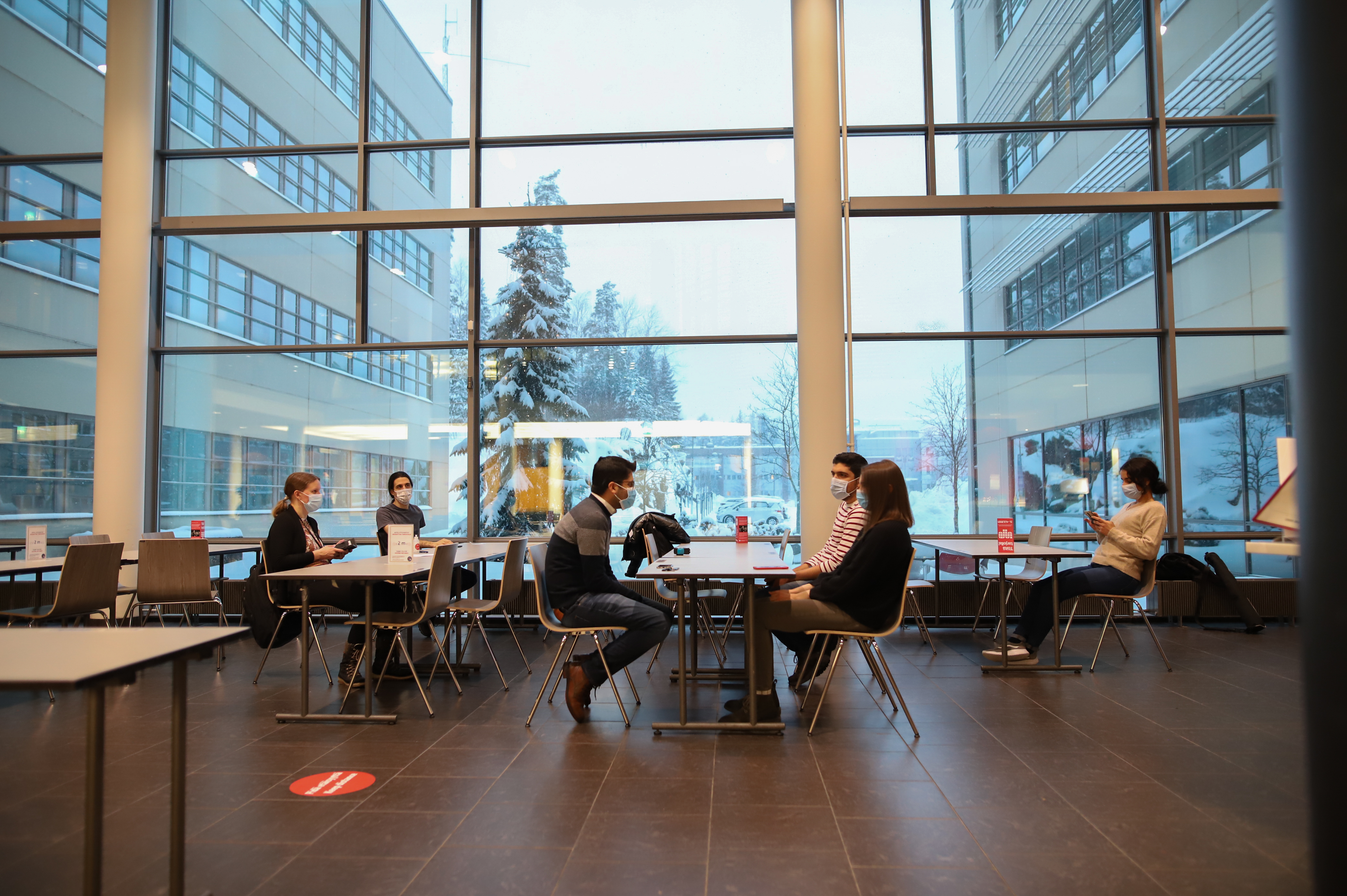}
    \includegraphics[width=0.24\linewidth]{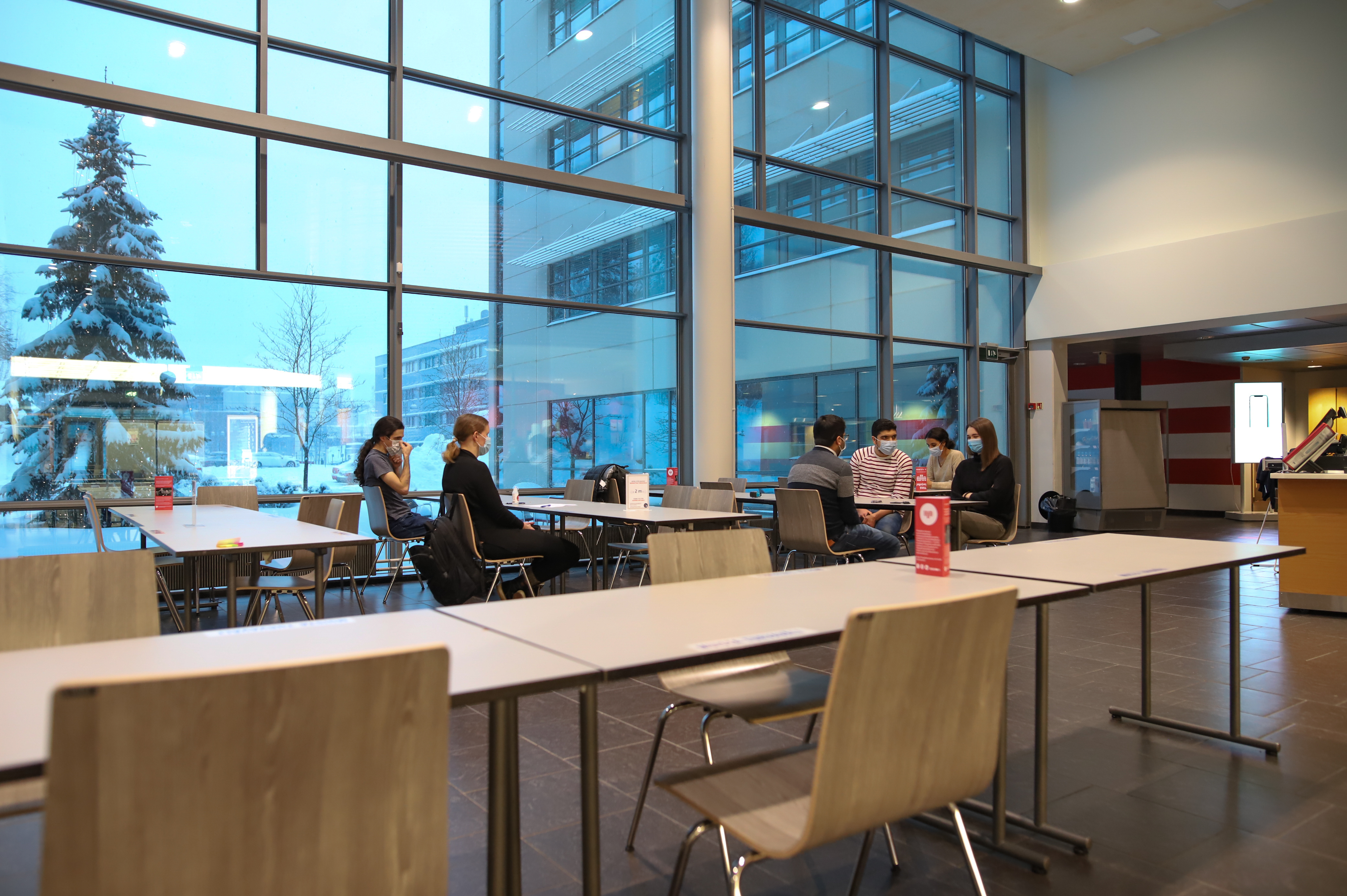}
    \includegraphics[width=0.24\linewidth]{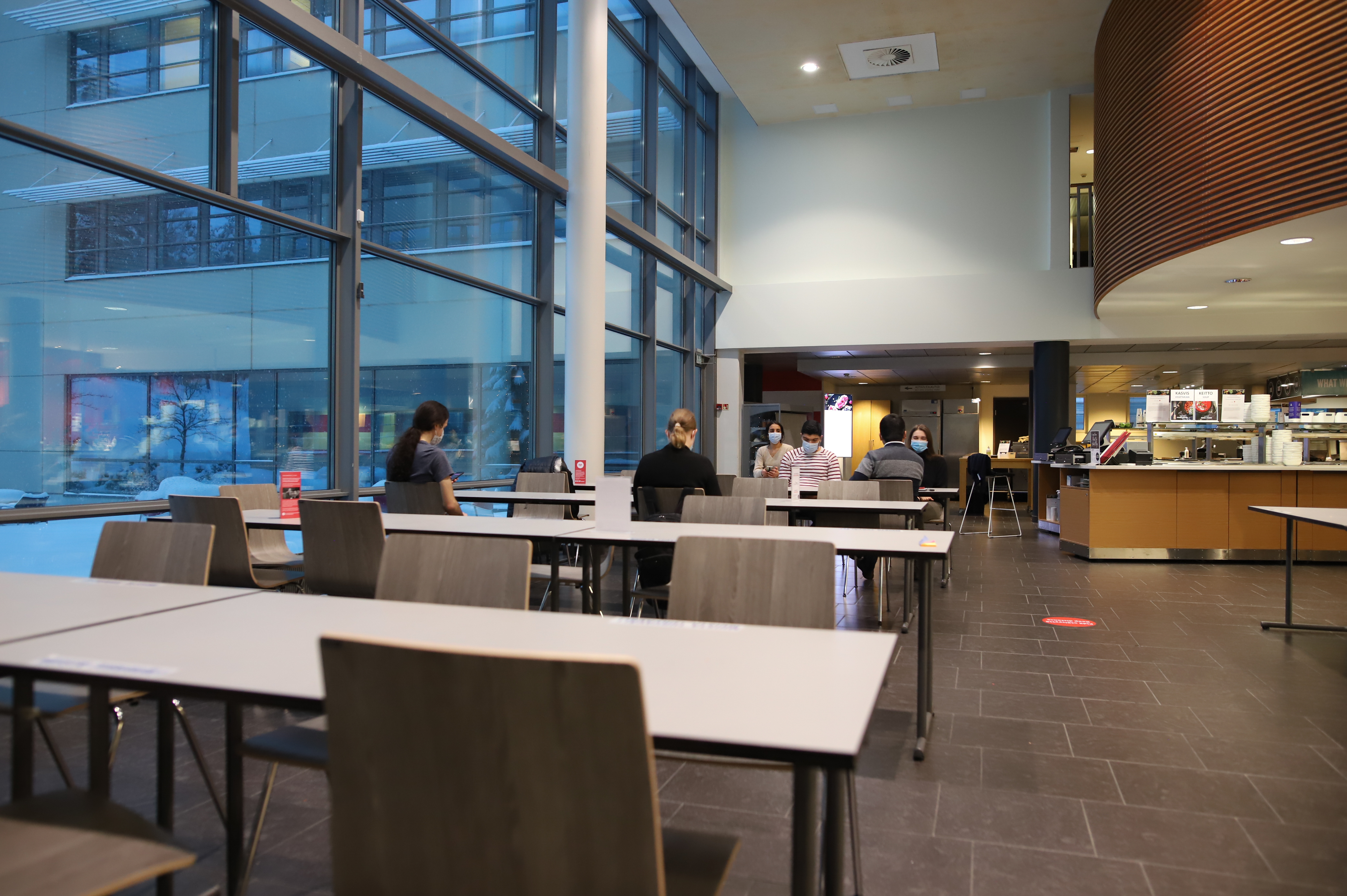}
    \caption{Example photos from the test dataset. The upper row has photos from the outdoor photo shoot taken from all camera positions C0 (left) to C2 (right) and the lower row has photos from the indoor photo shoot taken from all camera positions C0 (left) to C3 (right).}
    \label{fig:samplephotos}
\end{figure*}

The used camera models were Canon EOS 5D Mark II and Canon EOS 6D Mark II. The used focal length sizes were 16, 24, 35, 50, 105, 200, and 300 mm.  The cameras were stabilized on a tripod and the tripod's height was 135 cm for all images.  Fig.~\ref{fig:samplephotos} shows example photos from both photo shoots, one photo from each camera position.

\subsection{Test Data Description}
\label{ssec:datadescription}

The overall dataset contains 96 images including 63 outdoor images and 33 indoor images. All of the images are in JPG format. 81 of the images have the resolution 4180x2768 and 15 of the images have the resolution 4080x2720.  Two different camera models were used and the sensor size for both of these cameras is 36 mm in width and 24 mm in height. The distribution of the pictures in terms of focal lengths, camera models, and shooting setting are given in Table 1. 
\begin{table}[t]
\centering
\begin{tabular}{|c|c|cc|}
\hline
\multirow{2}{*}{\begin{tabular}[c]{@{}c@{}}Focal \\ Length (mm)\end{tabular}} & \multirow{2}{*}{Camera Model} &
  \multicolumn{2}{c|}{Shooting Setting} \\ \cline{3-4} 
 & &
  Indoor &
  Outdoor \\
 \hline
16  & Canon EOS 6D Mark II & 4  & 7   \\
24  & Canon EOS 6D Mark II & 4  & 8   \\
35  & Canon EOS 6D Mark II & 4  & 11  \\
50  & Canon EOS 6D Mark II & 7  & 11  \\
105 & Canon EOS 6D Mark II & 14 & 11  \\
200 & Canon EOS 5D Mark II & -  & 7   \\
300 & Canon EOS 5D Mark II &-  & 8   \\ \hline
All & &33  & 63   \\ \hline
\end{tabular}
\caption{Numbers of photos in the test dataset for different focal lengths (mm), camera models, and shooting settings (indoor/outdoor).}
\end{table}

\begin{figure*}
     \centering
     \begin{subfigure}[b]{0.6\textwidth}
         \centering
         \includegraphics[width=\textwidth]{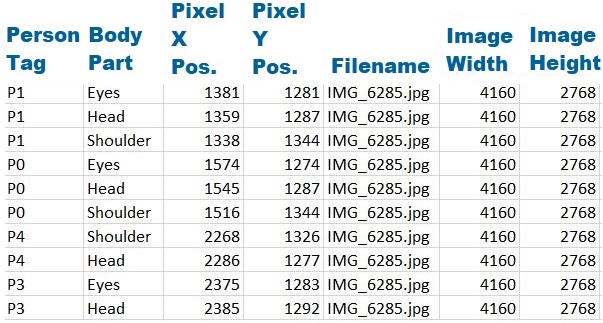}
         \caption{Body part pixel locations}
         \label{sfig:bodyparts}
     \end{subfigure}
     \\
     \begin{subfigure}[b]{0.43\textwidth}
         \centering
         \includegraphics[width=\textwidth]{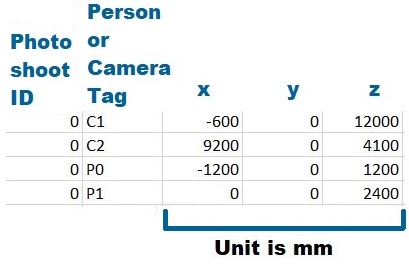}
         \caption{Ground truth relative 3D location}
         \label{sfig:3dlocations}
     \end{subfigure}
     \begin{subfigure}[b]{0.38\textwidth}
         \centering
         \includegraphics[width=\textwidth]{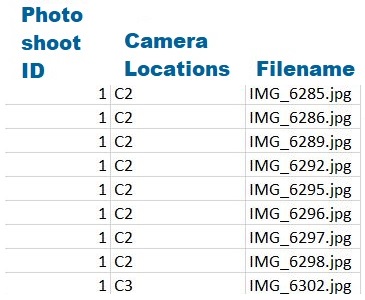}
         \caption{Photoshoot identifiers and camera locations}
         \label{sfig:photshoot}
     \end{subfigure}
        \caption{Annotation file formats}
        \label{fig:datasetexplanation}
\end{figure*}

Along with the images, we also provide different annotation data provided in three separate .csv files illustrated in Fig.~\ref{fig:datasetexplanation}. The first file (Fig.~\ref{sfig:bodyparts}) contains the pixel locations of four different body parts. These annotated body parts are center of the eyes, center of the shoulders, center of the torso, and center of the head. If a body part is not visible in the image, then it is not annotated. The people in the images are labeled as P0, P1, P2, P3, P4, P5, and P6 in the annotation file. These person tags are consistent through all of the images. This means that a person tag always refers to the same person in all of the images that we provide. The second file (Fig.~\ref{sfig:3dlocations}) contains the 3D locations of people and different camera positions in both photoshoots. Photoshoot ID 0 refers to the outdoor photo shoot and photoshoot ID 1 refers to the indoor photo shoot. The third file (Fig.~\ref{sfig:3dlocations}) links the image filenames with the corresponding photoshoot and camera location. The cameras' exterior orientation parameters are not included in the metadata of the images.

New images can be added to the dataset simply by following the described structure of the annotation data shown in Fig.~\ref{fig:datasetexplanation}. This does not require any changes in the provided evaluation codes. New photo shoots, i.e., new settings of people, must be identified with a unique integer identifier. For any photo shoot, the real world locations of the people should stay the same in all the photos. There may be pictures taken from different camera locations. Person and camera tags should start with a P and C letter, respectively, followed by a unique identifier integer. Person and camera location tags must be consistent within a given photo shoot, however repeated tags in different photo shoots are allowed. This means that two different people or camera tags could be the same as long as they belong to a different photo shoot. At least 1 of 4 body parts (center of the eyes, shoulders, torso, head) of the people in the images must be annotated in terms of pixel locations. They should be named "Eyes", "Shoulder", "Torso", and "Head" in the body part column of the body part pixel location file in Fig.~\ref{sfig:bodyparts}. 

To be consistent with the annotations in the provided test images, the annotation can be done as follows. Using the keypoint numbering in Fig.~\ref{fig:openposekeypoints}, the center of the eyes refers to the middle point of the keypoint pair 15-16, the center of the shoulders refers to the middle point of the keypoint pair 2-5, the center of the torso refers to the middle point of the keypoint pair 1-8, and the head should be annotated as middle point of the head regardless of the head's angle with respect to the camera. If a head is sideways and only one of the eyes is visible, the visible eye can be annotated as the center of the eyes. If none of the eyes are visible, the center of the eyes should not be annotated. The center of the eyes should also not be annotated if at least one of the eyes is out of the picture due to the head being on the edge of the picture. The other body parts can be annotated as long as they are either completely visible in the picture or are partially occluded by another person or object. In the cases where they are partially occluded, the pixel location should be estimated as if the occluding person or object was not present in the picture. The center of the shoulders, torso, and head should not be annotated only in the cases where these body parts are either partially or completely out of the picture due to the person being on the edge of the picture. If a person is sideways and only one of the shoulders, i.e., keypoints 2 and 5, is visible, this point can be annotated as the center of the shoulders.   \par

\subsection{Evaluation Protocol}
\label{ssec:protocol}
Any distance estimation method to be tested using the benchmark should give as output at least 1 of the 4 annotated pixel body locations along with either the estimated 3D location of the persons or the estimated distances between the people. The body part can be different for each person, or a method may choose to give only a single body part, such as the head, for all the persons. The test benchmark uses the pixel locations to automatically match each detected person with one of the ground truth locations and then computes average percentual pair-wise estimation errors between the estimated and ground truth distances. 

We provide all the necessary functionalities for testing as long as the required output for each image is given. Internally, the matching is carried out by comparing the automatically detected body pixel locations with the points annotated in the files. The automatically detected body parts are compared to all of the respective annotated body parts. As an example, a detected torso point is compared to all of the annotated torso points for that image. For all of the detected body parts of a person, the closest respective annotated point in terms of pixel-wise distance is found. In case there are more than one detected persons matched with the same ground truth person, the matching is done in a greedy manner by selecting only the closest match and the rest of the detected persons for that ground truth person are regarded as false positives.

After matching the detections with the persons labeled in the photos, we calculate the distances between each person pair by using their estimated 3D locations. Then, the estimated pair-wise distances are compared to the corresponding ground truth pair-wise distances to obtain a percentual distance estimation error for each pair. The performance is evaluated by taking the average of all of the pair-wise percentual distance estimation errors for each image and then averaging over images. In addition to the pair-wise percentual distance estimation error, we evaluate also the person detection rate, i.e., the ratio of correctly detected person averaged over all the images, and the false discovery rate averaged over all the images. It should be noted here that we do not use any threshold for matching the detections with the actual people. As long as \hl{the number of detections is lower or equal to the actual number of images, all the detections are matched. Thus, d}etections can be considered false positives only if there are more detections than actual people for an image. Therefore, a method producing many false positive detections is expected to get a high detection rate, but naturally the distance estimations would likely be poor and the false discovery rate would be higher. On the other hand, a method missing most the people could have a low pair-wise percentual distance estimation error for the detected people, but still not be suitable social distancing analyses. Therefore, it is important to consider all these metrics together, when evaluating a social distance estimation algorithm. 

The pair-wise percentual distance estimation error $D_{e}$ for \hl{the $e^{th}$} single image is given by the following formula where \hl{$n$} is the number of detected people in the image, \hl{$E_{i}$} is the estimated 3D location of the \hl{$i^{th}$} person and \hl{$G_{i}$} is the ground truth 3D location of the \hl{$i^{th}$} person:
\begin{equation}
D_{e} = \frac{\sum_{k = 1}^{n-1} \sum_{i = k+1}^{n} \frac{\left \|E_{k} - E_{i}\right \| - \left \|G_{k} - G_{i}\right \|}{\left \|G_{k} - G_{i}\right \|} * 100}{\binom{n}{2}}.
\end{equation}
Here, the distances may be also directly given instead of the 3D locations.

\begin{figure*}[t]
    \centering
    \includegraphics[width=0.494\linewidth]{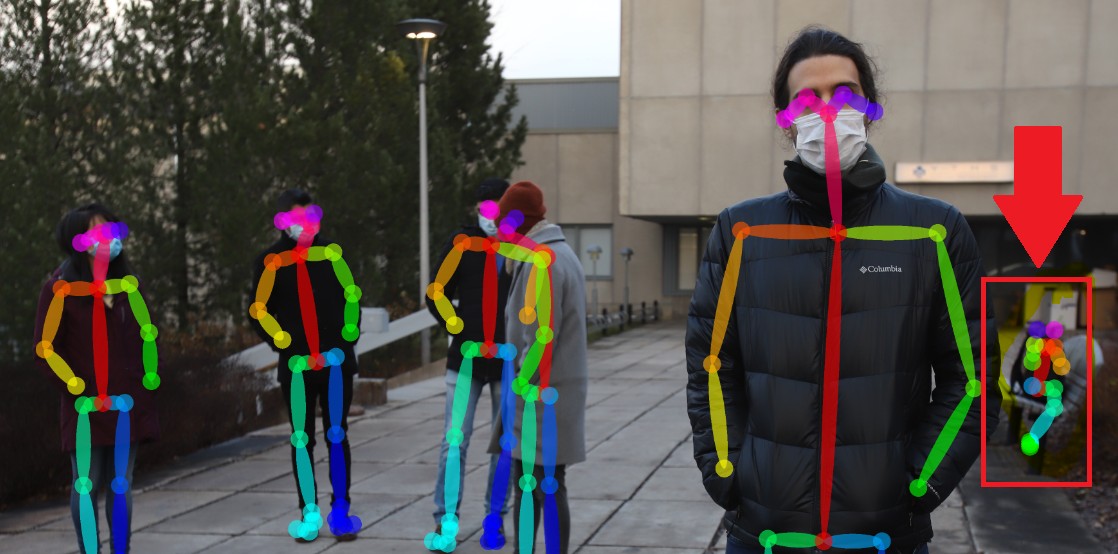}
    \includegraphics[width=0.494\linewidth]{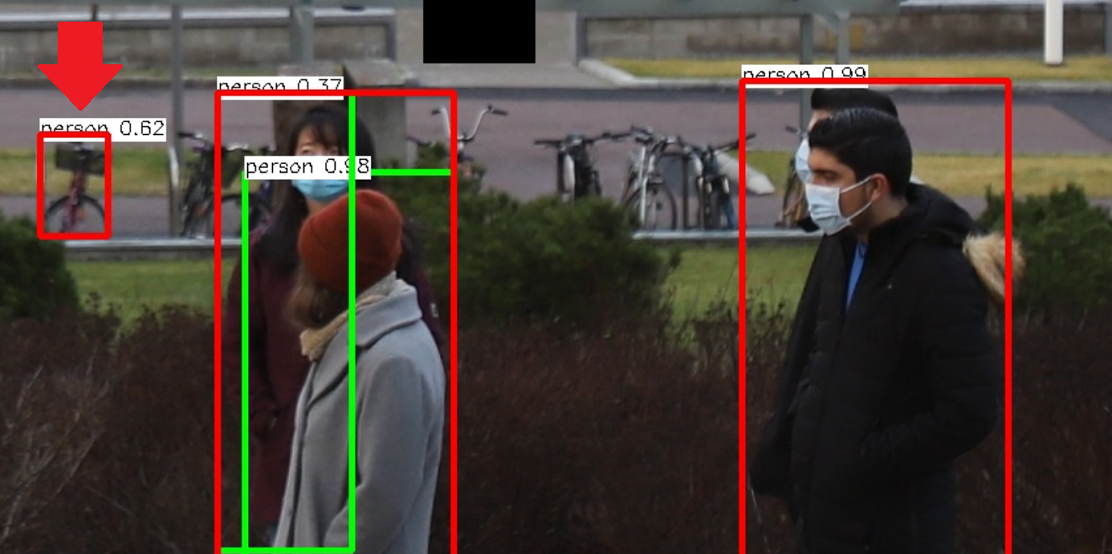}
    \caption{False positive examples for OpenPose (left) and YOLOv4 (right).}
    \label{fig:falsepositives}
\end{figure*}

In order to obtain an overall distance estimation error metric for a set of images, $D_{e}$ of all of the images in the image set are averaged. The distance estimation error for a set of images $D_{E}$ is given by the following formula where $N$ is the number of images in the set:\par
\begin{equation}
D_{E} = \frac{\sum_{e = 1}^{N} D_{e}}{N}.
\end{equation}

The test benchmark gives $D_{E}$,  the person detection rate, and the false discovery rate as an output for a given set of images as long as the input and annotated data are provided in the proper format. Currently, the test benchmark uses our provided test photos, but if new images are added to the dataset as explained in Section~\ref{ssec:datadescription}, these will be automatically considered in the evaluation.

\section{Proposed Method for Social Distance Estimation}
\label{sec:method}

Our proposed method to estimate social distances takes advantage of object detection and human pose estimation methods. Firstly, the input image is given to YOLOv4 \cite{yolov4} object detection model to obtain bounding boxes for people. After bounding boxes are obtained, overlapping boxes are grouped together. Then, these grouped boxes are cropped from the full image and they are individually given to OpenPose \cite{openpose1,openpose2,openpose3,openpose4} human pose estimation model. After the skeleton keypoints are extracted from OpenPose, the pixel locations of these keypoints are used in our distance estimation algorithm to obtain 3D location estimates for each person in the image. \par

When YOLOv4 and OpenPose models are used together, they eliminate the other model's false positives. \hl{The l}eft image in Fig.~\ref{fig:falsepositives} shows a case where a backpack is falsely recognized as a human by OpenPose. However, YOLOv4 does not recognize it as a human. Therefore, the backpack would not be cropped and given to the OpenPose model. \hl{The r}ight image in Fig.~\ref{fig:falsepositives} shows a case where a bicycle is falsely recognized as a human by the YOLOv4 model. The bicycle is then cropped from the full image and given to the OpenPose model. However, the OpenPose model does not detect any human skeleton in the cropped bicycle image. Therefore, neither of these false positive cases is further processed by the distance estimation algorithm. \par

\begin{figure}[!t]
    \centering
    \includegraphics[scale=0.5]{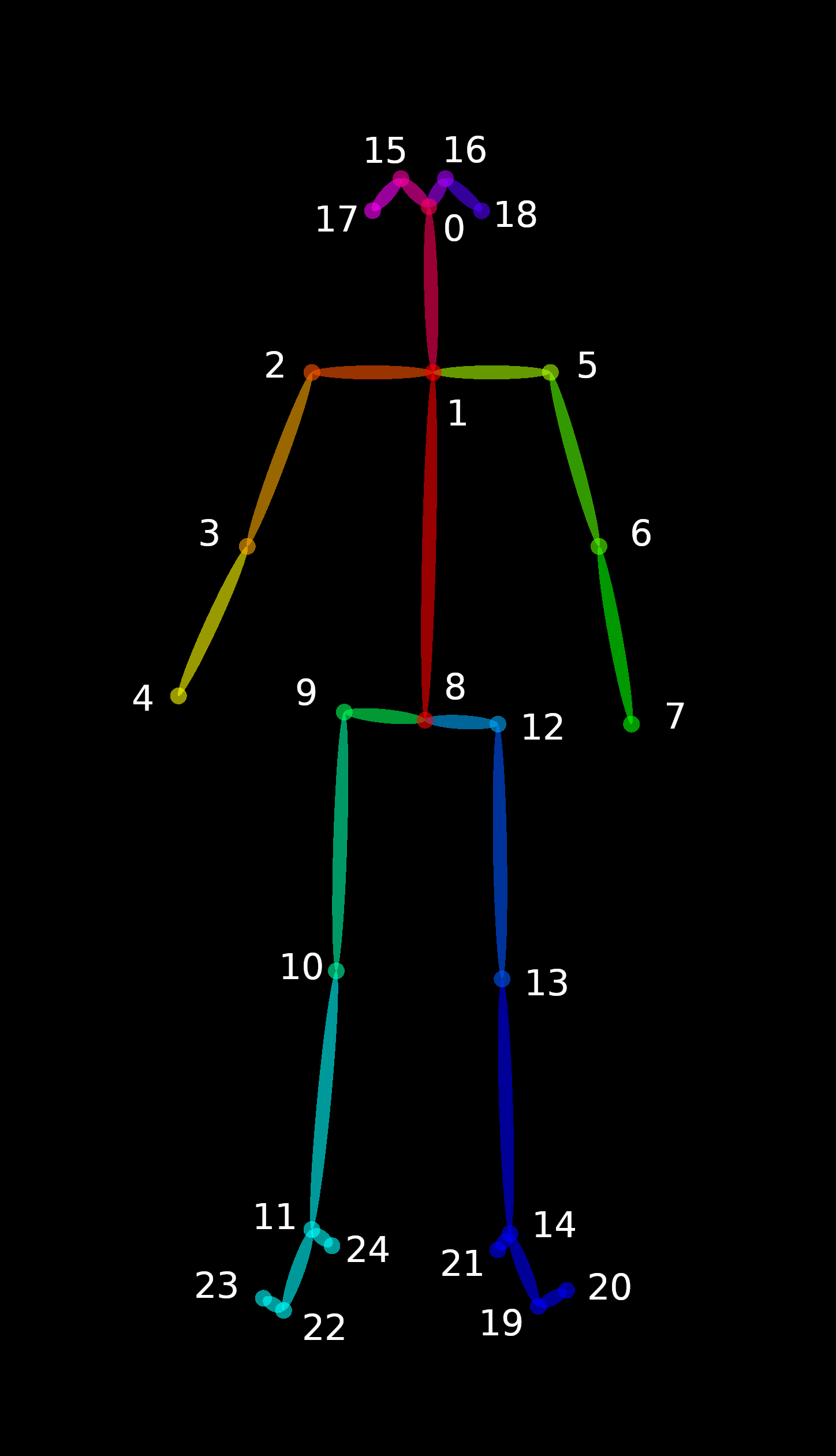}
    \caption{25 skeleton keypoint output of OpenPose.}
    \label{fig:openposekeypoints}
\end{figure}

After the cropped images from YOLOv4 are processed by the OpenPose model, the skeleton keypoints for detected human bodies are extracted. We use the 25 keypoint output version of OpenPose \hl{illustrated} in Fig.~\ref{fig:openposekeypoints}. \hl{Out of the extracted keypoints, we select pairs whose mutual distance is independent of the person's pose, whose average distance is available in the literature, whose angle towards the lens is as constant as possible, and which are visible in most of the photos. With these criteria, we select three key point pairs for our algorithm: 15-16  for pupillary distance, 2-5 for shoulder width, and 1-8 for torso length. In typical media or personal photos, the torso has the most constant angle towards the lens, but the eyes and shoulders are visible also in the close-up and portrait photos, where the torso is not seen. We assume average adult body proportions for the three keypoint pairs: 389 mm  for shoulder width \mbox{\cite{shoulderwidth}}, 63 mm for pupillary distance \mbox{\cite{pupildistance}} and 444 mm for torso length \mbox{\cite{torsolength}}.} The extracted keypoint \hl{pairs} are then processed by our distance estimation algorithm that estimates 3D positions with respect to the camera for each person. 

\begin{figure*}[h]
    \centering
    \includegraphics[width=0.85\linewidth]{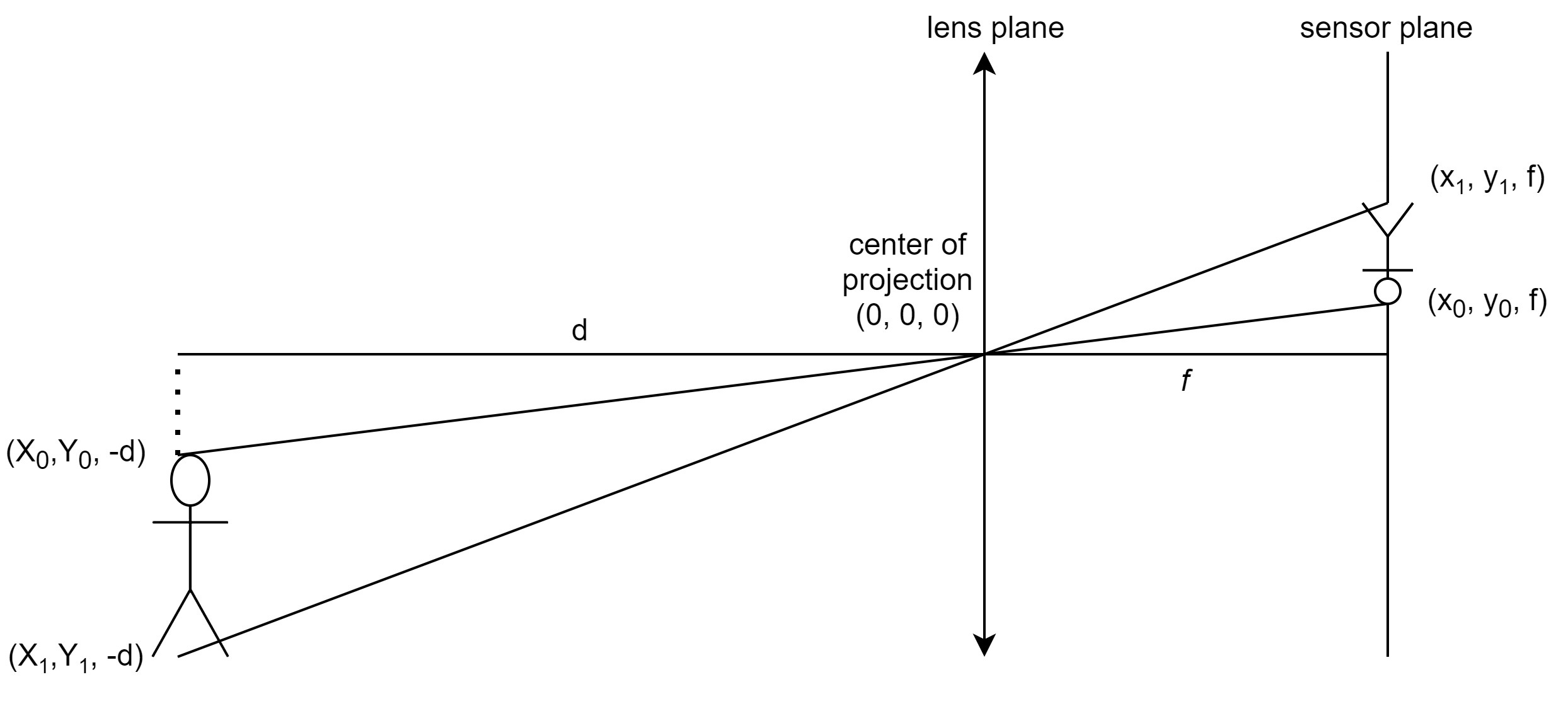}
    \caption{Pinhole camera model.}
    \label{fig:pinhole}
\end{figure*}

\begin{figure}[!t]
    \centering
    \includegraphics[width=1\linewidth]{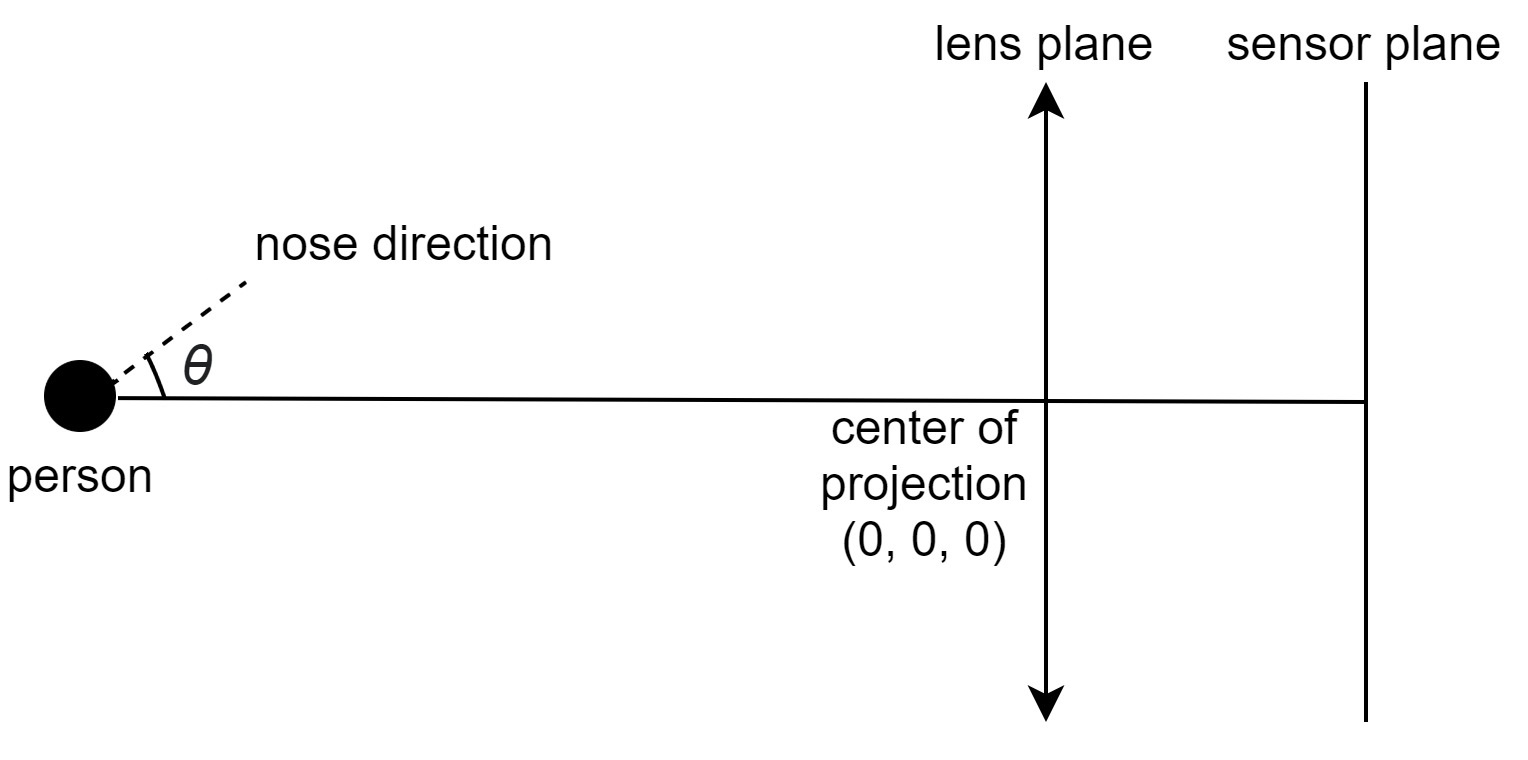}
    \caption{Birdseye view of orientation angle toward the lens.}
    \label{fig:noseangle}
\end{figure}

We use the pinhole camera model \cite{pinhole} shown in Fig.~\ref{fig:pinhole} for our calculations. We also make an assumption that every \hl{keypoint pair} is parallel to the camera's sensor plane. We make th\hl{ese} assumption\hl{s} because the \hl{subjects' poses and} camera's exterior orientation parameters \cite{camparameters} are not known. Estimating the exterior orientation parameters \cite{camparameters} of the camera from single images is an ill-posed problem \cite{illposed}, \hl{but} in most cases the angle between a person's torso and the camera's sensor plane is negligible for our calculations.

We denote 3D locations of the keypoints on the image coordinate system as
\begin{equation}
(x_{a}, y_{a}, f),
\label{eq:icoordinate}
\end{equation}
\hl{where $f$ is the focal length,}
and 3D location estimates of the keypoints on the world coordinate system as
\begin{equation}
E_{n} = (X_{a}, Y_{a}, -d),
\label{eq:wcoordinate}
\end{equation}
\hl{where $d$ is the distance to the camera.} 
The distance between a pair of keypoints on the image coordinate system is
\begin{equation}
D_{i} = \sqrt{(x_{0}-x_{1})^{2} + (y_{0}-y_{1})^{2} + (f-f)^{2}}
\label{eq:idistance}
\end{equation}
and the distance between the keypoints on the world coordinate system is
\begin{equation}
D_{w} = \sqrt{(X_{0}-X_{1})^{2} + (Y_{0}-Y_{1})^{2} + (d-d)^{2}}.
\label{eq:wdistance}
\end{equation}

Since the camera sensor's plane size is known, $x_{a}$ and $y_{a}$ in Eq.~\eqref{eq:icoordinate} \hl{can be}  derived from the the x and y pixel locations of the keypoints in the image. The last coordinate, $f$, in Eq.~\eqref{eq:icoordinate} is obtained from the camera parameters. Thus, all the keypoints' 3D positions on the image coordinate system in Eq.~\eqref{eq:icoordinate} are known and $D_i$ can be solved. By using triangle similarity, the following equations give 3D positions of the keypoints on the world coordinate system. Eq.~\eqref{eq:destimate}, where $D_w$ is one of the average body proportions, is used to derive $d$ in Eq.~\eqref{eq:wcoordinate}. After $d$ is derived, $X_{a}$ and $Y_{a}$ are obtained from Eqs.~\eqref{eq:Xa} and \eqref{eq:Ya}.

\begin{equation}
\frac{D_{i}}{f} = \frac{D_{w}}{d}
\label{eq:destimate}
\end{equation}

\begin{equation}
X_{a} = -\frac{d}{f} x_{a}
\label{eq:Xa}
\end{equation}

\begin{equation}
Y_{a} = -\frac{d}{f} y_{a}
\label{eq:Ya}
\end{equation}

\begin{figure*}[t]
    \centering
    \includegraphics[width=0.32\linewidth]{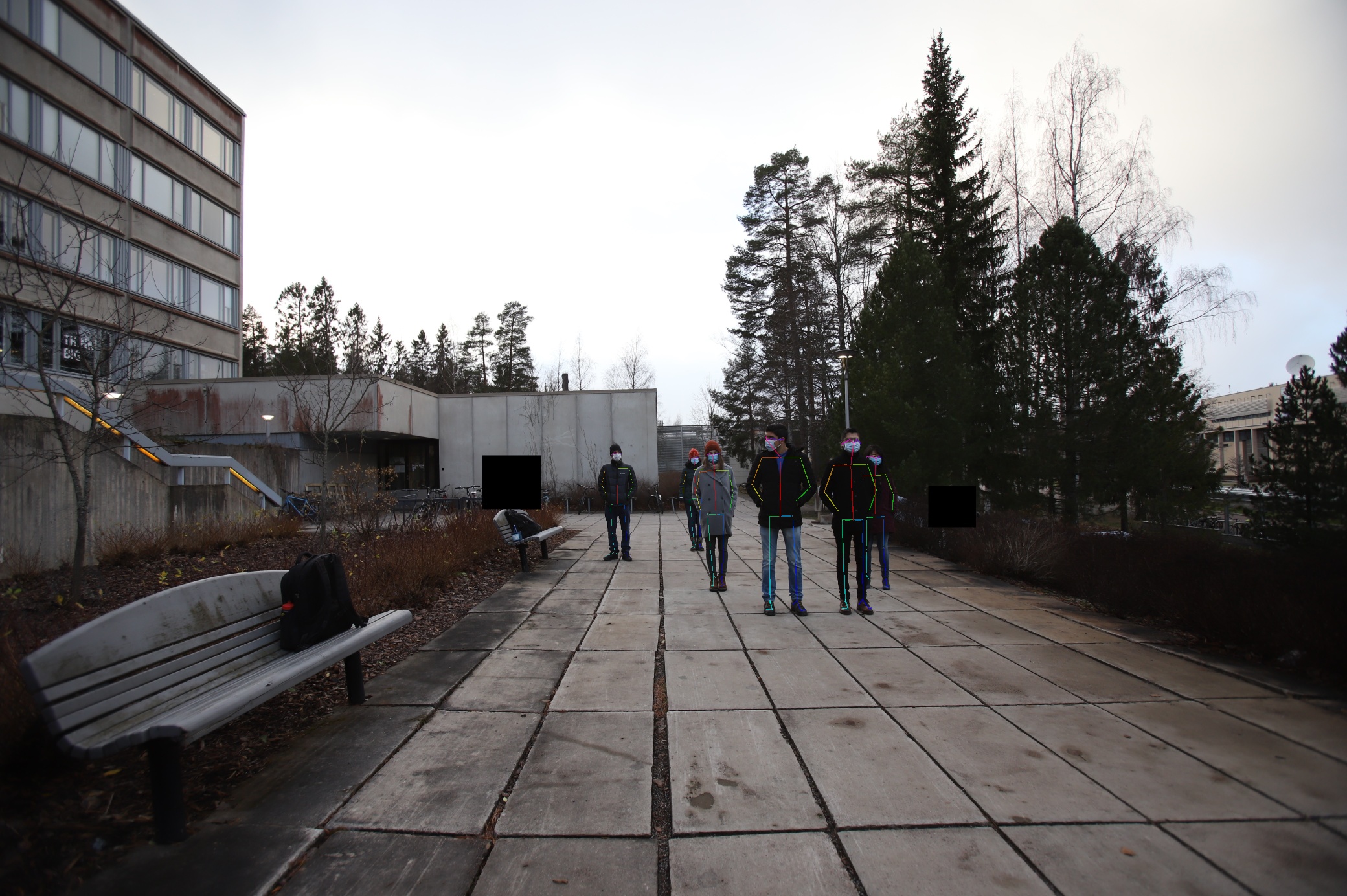}
    \includegraphics[width=0.32\linewidth]{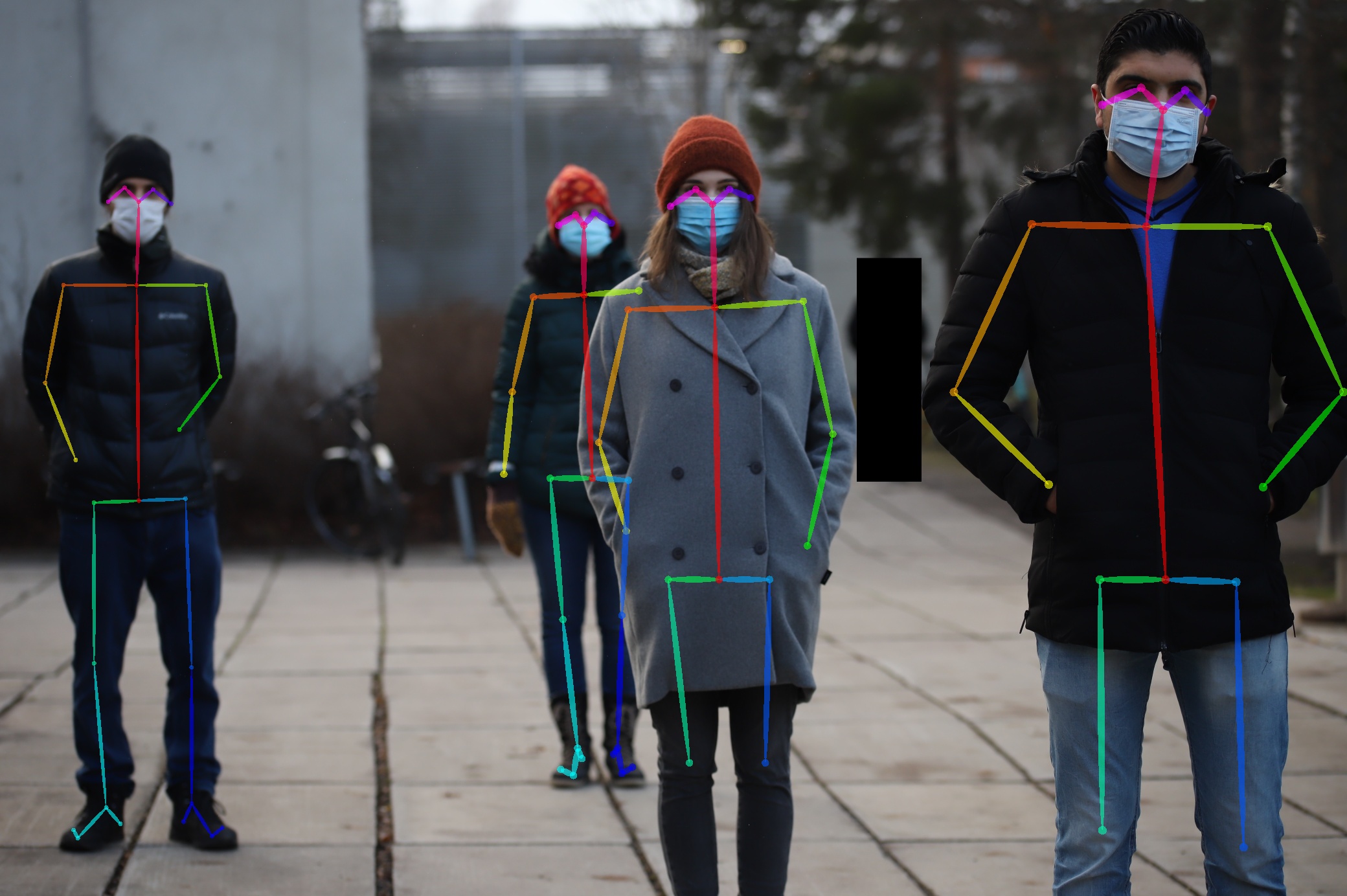}
    \includegraphics[width=0.32\linewidth]{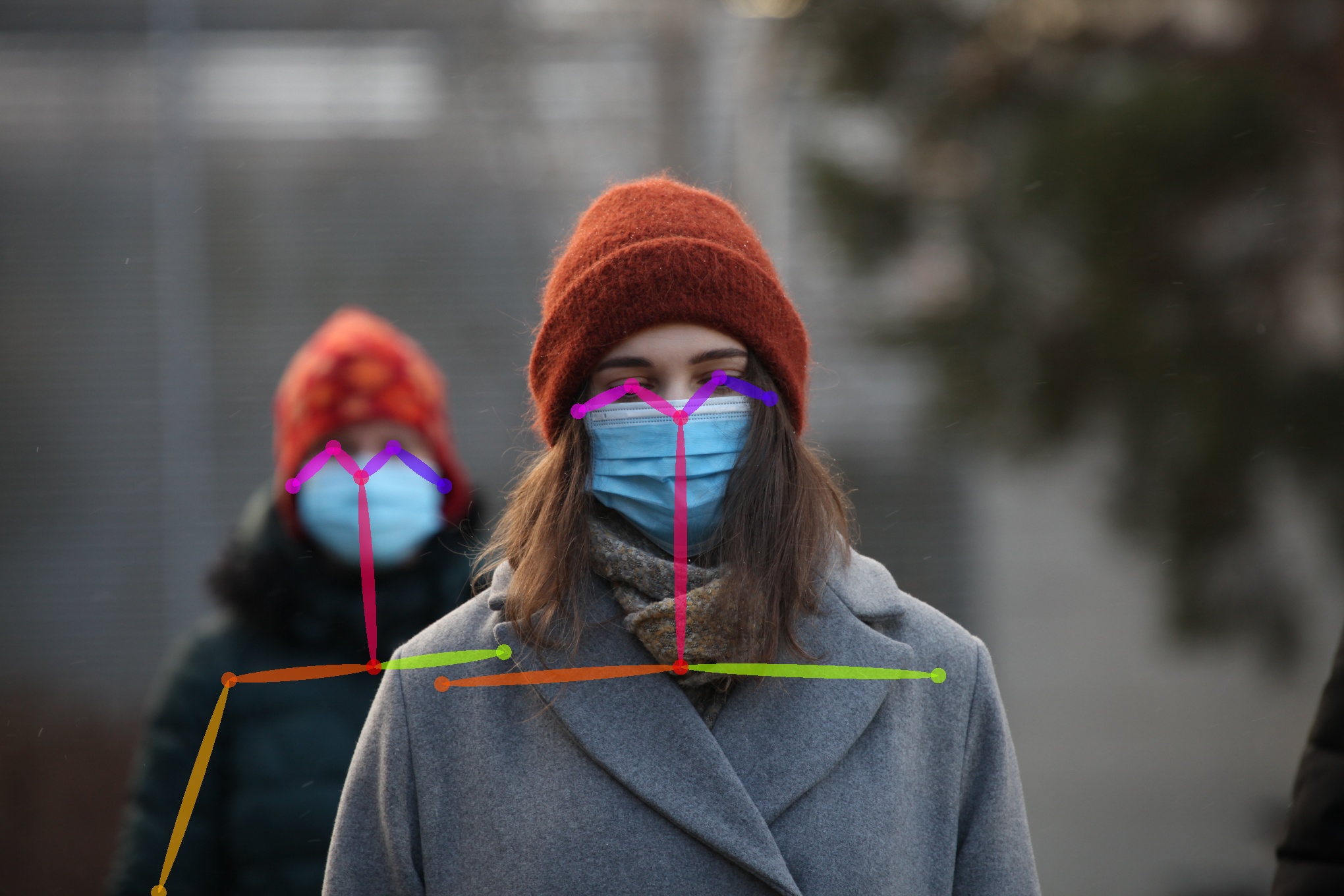}
    \caption{Examples of pictures from the dataset belonging to the first photo shoot, all of them taken from camera location C1. The used focal lengths for the pictures are 16mm, 105mm and 300mm from left to right.}
    \label{fig:focallengths}
\end{figure*}

\begin{figure}[p]
    \centering
    \includegraphics[scale=0.13]{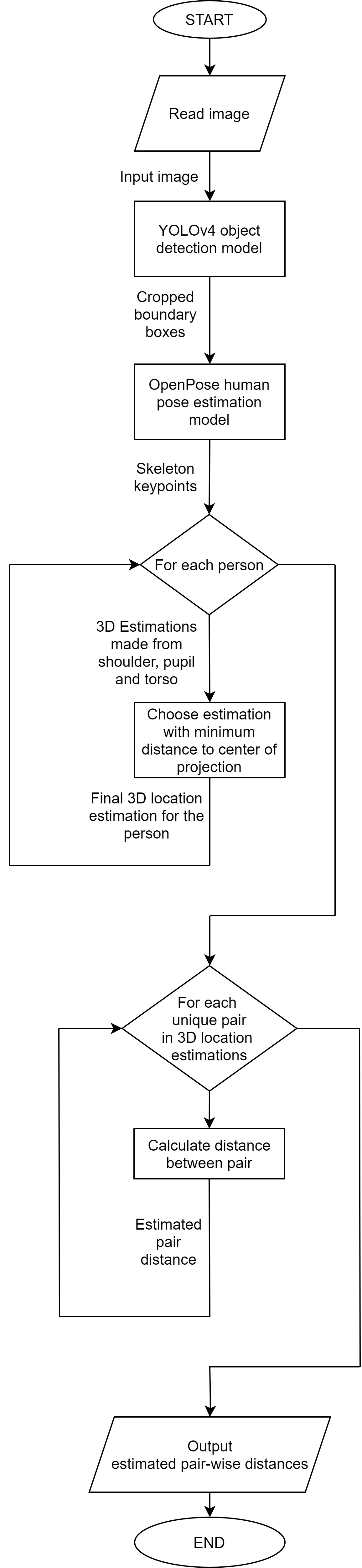}
    \caption{Flowchart of the method.}
    \label{fig:flowchart}
\end{figure}

After the 3D coordinates of the keypoints on the world coordinate system in Eq.~\eqref{eq:wcoordinate} are estimated, the middle points of each detected keypoint pair are used to represent a 3D location for the person. Thus, we have at most 3 different estimated 3D locations for a person, one for each keypoint pair (shoulder, pupil, torso). \hl{While we assume that the keypoint pairs are parallel to the camera's sensor plane, this assumption may not be valid, and the accuracy of the estimated locations is affected by the severity of the violations.}
Fig.~\ref{fig:noseangle} shows the birdseye view of a person's orientation angle $\theta$ toward the lens. \hl{If the angle is non-zero,} 
the shoulder and pupil keypoint pairs are \hl{ no longer parallel to the sensor plane and the estimates based on these keypoint pairs are} prone to error. However, \hl{in a typical situation of upright torsos} the estimates made from the torso length are unaffected by $\theta$, \hl{because} $\theta$ does not affect $D_{i}$ computed using Eq.~\eqref{eq:idistance} for the torso. \hl{On the other hand, also a torso may not be parallel to the sensor plane either because the person is in a bent position or because the camera's pitch angle is non-zero. For an overhead image, shoulders might be parallel to the sensor plane, while torsos would be perpendicular. Whenever the assumption on a keypoint pair being parallel to the sensor plane is violated,} $D_{i}$ in Eq.~\eqref{eq:idistance} decreases. A smaller $D_{i}$ leads to a larger estimate for $d$ from Eq.~\eqref{eq:destimate}. For this reason, we select the 3D location with the smallest distance to the camera. \hl{For typical media or personal photos, where the pitch angle is small, this} means using the estimate derived from the torso whenever it is available. However, for close-up and portrait pictures, the torso is often not visible.  Fig.~\ref{fig:focallengths} shows three pictures taken from the same location but with increasing focal lengths. The rightmost image in Fig.~\ref{fig:focallengths} is an example of a close-up picture where the distance estimations have to be made from the shoulder and pupil distances since there are no visible torsos. 

Finally, our method computes the distances between all the pairs of detected people and gives them as outputs. The pixel locations for the detected persons are given to be able to evaluate on our benchmark, while they are not needed if the method is used for analysing social distancing in novel images. The overall flowchart of the proposed social distance estimation method is illustrated in Fig.~\ref{fig:flowchart}.

\section{Experimental Results}
\label{sec:experimental}

\subsection{Experimental Setup}
All of the code was developed in Python programming language version 3.8 \cite{van1995python}. OpenPose \cite{openpose1,openpose2,openpose3,openpose4} and YOLOv4 \cite{yolov4} models were used for human detection. All of the images were resized to 50\% of their original sizes before being sent to YOLOv4 model. The input size of YOLOv4 was set to 704x704. Input size was not set for OpenPose as OpenPose is able to adapt its input size for each image. The version of the OpenPose model we were using was originally trained by using the COCO keypoint challenge dataset \cite{COCO}, combined with OpenPose authors' own annotated dataset for foot keypoint estimation which consists of a small subset of the COCO dataset where the authors labelled foot keypoints. YOLOv4 uses CSPDarknet53 \cite{darknet} as its backbone which was trained on the ImageNet dataset \cite{imagenet}. The deep learning models were downloaded from their respective official source code pages \footnote{https://github.com/CMU-Perceptual-Computing-Lab/openpose} \footnote{https://github.com/AlexeyAB/darknet} and they were loaded and used by TensorFlow library version 2.3.1 \cite{tensorflow2015}. For image processing purposes, OpenCV imaging library was used \cite{opencv}.  In addition to our final method that generates 3D position estimates using torso, shoulders, and eyes and selects the estimate closest to the camera as explained in Section~\ref{sec:method}, we also evaluate variants of the proposed method, where only one of these body parts is used at the time.  We use our test benchmark to compute the results for all the images and for outdoor and indoor images separately.

\subsection{Results}

\begin{table*}[]
\centering
\begin{tabular}{|c|c|cc|cc|cc|cc|}
\hline
\multirow{2}{*}{\begin{tabular}[c]{@{}c@{}}Focal \\ Length (mm)\end{tabular}} & \multirow{2}{*}{\begin{tabular}[c]{@{}c@{}}Number \\ of \\ Pictures\end{tabular}} & \multicolumn{2}{c|}{\begin{tabular}[c]{@{}c@{}}Shoulder\\  Based \\ Method\end{tabular}}                                                                                & \multicolumn{2}{c|}{\begin{tabular}[c]{@{}c@{}}Pupil \\ Based\\ Method\end{tabular}}                                                                                    & \multicolumn{2}{c|}{\begin{tabular}[c]{@{}c@{}}Torso \\ Based\\ Method\end{tabular}}                                                                                    & \multicolumn{2}{c|}{\begin{tabular}[c]{@{}c@{}}Combined\\ Method\end{tabular}}                                                                                          \\ \cline{3-10} 
                                                                              &                                                                                   & \multicolumn{1}{c|}{\begin{tabular}[c]{@{}c@{}}Person\\ Detection\\ Rate\end{tabular}} & \begin{tabular}[c]{@{}c@{}}Pair-wise\\ Percent\\ Distance\\ Error\end{tabular} & \multicolumn{1}{c|}{\begin{tabular}[c]{@{}c@{}}Person\\ Detection\\ Rate\end{tabular}} & \begin{tabular}[c]{@{}c@{}}Pair-wise\\ Percent\\ Distance\\ Error\end{tabular} & \multicolumn{1}{c|}{\begin{tabular}[c]{@{}c@{}}Person\\ Detection\\ Rate\end{tabular}} & \begin{tabular}[c]{@{}c@{}}Pair-wise\\ Percent\\ Distance\\ Error\end{tabular} & \multicolumn{1}{c|}{\begin{tabular}[c]{@{}c@{}}Person\\ Detection\\ Rate\end{tabular}} & \begin{tabular}[c]{@{}c@{}}Pair-wise\\ Percent\\ Distance\\ Error\end{tabular} \\ \hline
16   & 11   & 0.84    & 163.91     & 0.68          & 45.81   & 0.77   & \textbf{19.94}     & \textbf{0.90}        & 20.50                                                                          \\
24      & 12   & 0.83    & 230.80     & 0.61    & 65.20     & 0.84   & \textbf{17.76}   & \textbf{0.93}    & 21.62                                                                          \\35      & 15     & 0.84   & 101.48   & 0.70      & 52.86  & \textbf{0.96}    & \textbf{20.51}     & \textbf{0.96}     & 22.61                                                                          \\
50    & 18   & 0.84     & 190.37    & 0.62   & 50.35  & 0.84     & 26.16  & \textbf{0.91}   & \textbf{25.52}                                                                          \\
105     & 25   & 0.82     & 116.52      & 0.80    & 52.65     & 0.72    & 34.54     & \textbf{0.95}     & \textbf{34.15}    
\\
200     & 7   & 0.71    & 109.99     & 0.75     & \textbf{50.61}      & 0.41   & 93.54       & \textbf{0.79}     & 53.68                                                                          \\
300     & 8  & 0.72   & 288.13   & \textbf{0.89}    & \textbf{34.48}    & 0.18     & -   & \textbf{0.89}    & \textbf{34.48}    \\ \hline
All                                              & 96        & 0.81      & 166.45    & 0.72     & 51.59     & 0.73     & \textbf{27.55}      & \textbf{0.92}   & 28.88                                                                          \\ \hline
\end{tabular}
\caption{Person detection rates and pair-wise percentual distance errors for each of the methods for both of the photo shoots (indoor and outdoor) combined}
\label{tbl:combined}
\end{table*}

\begin{table*}[]
\centering
\begin{tabular}{|c|c|cc|cc|cc|cc|}
\hline
\multirow{2}{*}{\begin{tabular}[c]{@{}c@{}}Focal \\ Length (mm)\end{tabular}} & \multirow{2}{*}{\begin{tabular}[c]{@{}c@{}}Number \\ of \\ Pictures\end{tabular}} & \multicolumn{2}{c|}{\begin{tabular}[c]{@{}c@{}}Shoulder\\  Based \\ Method\end{tabular}}                                                                                & \multicolumn{2}{c|}{\begin{tabular}[c]{@{}c@{}}Pupil \\ Based\\ Method\end{tabular}}                                                                                    & \multicolumn{2}{c|}{\begin{tabular}[c]{@{}c@{}}Torso \\ Based\\ Method\end{tabular}}                                                                                    & \multicolumn{2}{c|}{\begin{tabular}[c]{@{}c@{}}Combined\\ Method\end{tabular}}                                                                                          \\ \cline{3-10} 
                                                                              &                                                                                   & \multicolumn{1}{c|}{\begin{tabular}[c]{@{}c@{}}Person\\ Detection\\ Rate\end{tabular}} & \begin{tabular}[c]{@{}c@{}}Pair-wise\\ Percent\\ Distance\\ Error\end{tabular} & \multicolumn{1}{c|}{\begin{tabular}[c]{@{}c@{}}Person\\ Detection\\ Rate\end{tabular}} & \begin{tabular}[c]{@{}c@{}}Pair-wise\\ Percent\\ Distance\\ Error\end{tabular} & \multicolumn{1}{c|}{\begin{tabular}[c]{@{}c@{}}Person\\ Detection\\ Rate\end{tabular}} & \begin{tabular}[c]{@{}c@{}}Pair-wise\\ Percent\\ Distance\\ Error\end{tabular} & \multicolumn{1}{c|}{\begin{tabular}[c]{@{}c@{}}Person\\ Detection\\ Rate\end{tabular}} & \begin{tabular}[c]{@{}c@{}}Pair-wise\\ Percent\\ Distance\\ Error\end{tabular} \\ \hline
16      & 7  & \textbf{0.85}    & 129.35    & 0.71    & 56.42    & \textbf{0.85}    & 18.75     & \textbf{0.85}      & \textbf{18.47}      \\
24      & 8 & 0.83    & 169.08     & 0.64  & 84.67   & \textbf{0.91}      & \textbf{16.98}      & \textbf{0.91}      & 20.71                                                \\
35      & 11   & 0.90     & 174.70     & 0.76    & 64.18     & \textbf{0.96}     & \textbf{20.16}     & \textbf{0.96}     & 21.08                                                 \\
50    & 11     & 0.88    & 192.92    & 0.70    & 65.19      & 0.89   & \textbf{24.33}     & \textbf{0.92}   & 26.40      \\

105    & 11      & \textbf{1.00}     & 127.59    & \textbf{1.00}     & 48.99     & 0.81      & 41.64    & \textbf{1.00}     & \textbf{33.09}    \\
200     & 7  & 0.71    & 109.99    & 0.75    & \textbf{50.61}     & 0.41   & 93.54    & \textbf{0.79}   & 53.68   \\
300     & 8    & 0.72    & 288.13   & \textbf{0.89}        & \textbf{34.48}    & 0.18    & -     & \textbf{0.89}     & \textbf{34.48}                                          \\ \hline
All                                              & 63      & 0.86   & 164.43    & 0.78   & 58.73      & 0.74     & \textbf{28.82}     & \textbf{0.91}   & 28.87    \\ \hline
\end{tabular}
\caption{Person detection rates and pair-wise percentual distance errors for each of the methods for the first photo shoot (outdoor) where every person is standing up}
\label{tbl:outdoor}
\end{table*}

\begin{table*}[]
\centering
\begin{tabular}{|c|c|cc|cc|cc|cc|}
\hline
\multirow{2}{*}{\begin{tabular}[c]{@{}c@{}}Focal \\ Length (mm)\end{tabular}} & \multirow{2}{*}{\begin{tabular}[c]{@{}c@{}}Number \\ of \\ Pictures\end{tabular}} & \multicolumn{2}{c|}{\begin{tabular}[c]{@{}c@{}}Shoulder\\  Based \\ Method\end{tabular}}                                                                                & \multicolumn{2}{c|}{\begin{tabular}[c]{@{}c@{}}Pupil \\ Based\\ Method\end{tabular}}                                                                                    & \multicolumn{2}{c|}{\begin{tabular}[c]{@{}c@{}}Torso \\ Based\\ Method\end{tabular}}                                                                                    & \multicolumn{2}{c|}{\begin{tabular}[c]{@{}c@{}}Combined\\ Method\end{tabular}}                                                                                          \\ \cline{3-10} 
                                                                              &                                                                                   & \multicolumn{1}{c|}{\begin{tabular}[c]{@{}c@{}}Person\\ Detection\\ Rate\end{tabular}} & \begin{tabular}[c]{@{}c@{}}Pair-wise\\ Percent\\ Distance\\ Error\end{tabular} & \multicolumn{1}{c|}{\begin{tabular}[c]{@{}c@{}}Person\\ Detection\\ Rate\end{tabular}} & \begin{tabular}[c]{@{}c@{}}Pair-wise\\ Percent\\ Distance\\ Error\end{tabular} & \multicolumn{1}{c|}{\begin{tabular}[c]{@{}c@{}}Person\\ Detection\\ Rate\end{tabular}} & \begin{tabular}[c]{@{}c@{}}Pair-wise\\ Percent\\ Distance\\ Error\end{tabular} & \multicolumn{1}{c|}{\begin{tabular}[c]{@{}c@{}}Person\\ Detection\\ Rate\end{tabular}} & \begin{tabular}[c]{@{}c@{}}Pair-wise\\ Percent\\ Distance\\ Error\end{tabular} \\ \hline
16      & 4   & 0.83     & 224.38      & 0.62     & 29.89     & 0.62    & \textbf{22.71}      & \textbf{1.00}      & 24.05       \\
24       & 4   & 0.83    & 354.25     & 0.54       & 26.27    & 0.70      & \textbf{19.31}     & \textbf{0.95}     & 23.45                                                                          \\
35     & 4       & 0.66     & 51.94     & 0.54     & 24.56    & \textbf{0.95}    & 21.46     & \textbf{0.95}       & 26.82                                                  \\
50     & 7     & 0.76     & 186.35       & 0.50     & 31.27       & 0.76        & 29.03    & \textbf{0.88}        & \textbf{24.13}      \\
105       & 14    & 0.67    & 105.46     & 0.65      & 55.39      & 0.64      & \textbf{27.44}       & \textbf{0.91}    & 34.83        \\ \hline
All         & 33       & 0.73    & 170.57    & 0.59      & 38.92   & 0.71    & \textbf{25.24}     & \textbf{0.92}      & 28.90                                                                          \\ \hline
\end{tabular}
\caption{Person detection rates and pair-wise percentual distance errors for each of the methods for the second photo shoot (indoor) where every person is sitting down}
\label{tbl:indoor}
\end{table*}

Tables \ref{tbl:combined}, \ref{tbl:outdoor}, and \ref{tbl:indoor} show the person detection rates and pair-wise percentual distance estimation errors  for all the images, outdoor images, and indoor image, respectively. \hl{Since we are using YOLOv4 \mbox{\cite{yolov4}} in addition to OpenPose \mbox{\cite{openpose1,openpose2,openpose3,openpose4}} and they cancel each other's false positives, we have no cases with more detections than actual people in an image. This leads to zero false discovery rates as explained in Section \mbox{\ref{ssec:protocol}}}. Therefore, false discovery rates are not reported in the tables.
It can be observed from Table~\ref{tbl:combined} that the most reliable body part to estimate locations is the torso. However, estimations made from the torso alone fail for close-up pictures where the torso detection rate is low. When all three body parts (shoulder, pupil, and torso) are used together for the estimations, the obtained results shown in the last column are better than the results obtained from a single body part. Although the combined method has slightly lower performance than the torso based method in terms of pair-wise percent distance error, person detection rate is significantly higher than that of the torso based method.  The combined method mostly uses the torso whenever it is visible (overall shots) and uses the shoulder and pupil distances when the torso is not visible (close-up shots). \par
Looking at Tables \ref{tbl:outdoor} and \ref{tbl:indoor}, it can be seen that there are no significant differences in terms of pair-wise distance errors when it comes to indoor and outdoor pictures. However, it should be noted that the person detection rates for the outdoor pictures are higher than for the indoor pictures. This is primarily caused by the fact that the body parts of the people in the indoor pictures were obstructed by the chairs. There were also more cases of people facing away from the camera and people standing in front of other people in the indoor photo shoot.

\begin{table}[t]
\centering
\begin{tabular}{|c|c|c|}
\hline
\multirow{2}{*}{\begin{tabular}[c]{@{}c@{}}Number\\ of\\ Pictures\end{tabular}} & \multicolumn{2}{c|}{\begin{tabular}[c]{@{}c@{}}Combined\\ Method\end{tabular}} \\ \cline{2-3} 
 & \begin{tabular}[c]{@{}c@{}}Person\\ Detection\\ Rate\end{tabular} & \begin{tabular}[c]{@{}c@{}}Pair-wise\\ Percent\\ Distance\\ Error\end{tabular} \\ \hline
16 & 1.00 & 32.88 \\ \hline
\end{tabular}
\caption{\hl{Person detection rates and pair-wise percentual distance errors for the combined method for the photos taken from camera location C2, for which the zero pitch angle assumption is not valid.}}
\label{tbl:balcony}
\end{table}

\subsection{Additional results and analysis}
\label{sec:additionalresults}

\begin{figure}[h]
    \centering
    \includegraphics[scale=0.5]{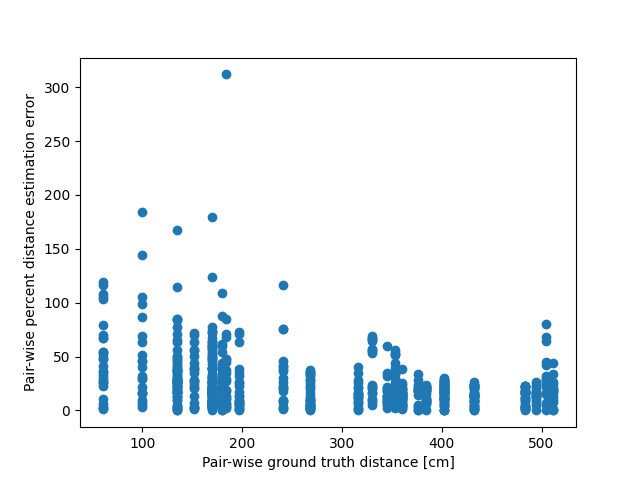}
    \caption{Pair-wise distance estimation errors for each of the ground truth pair-wise distances.}
    \label{fig:errorsgraph}
\end{figure}

\hl{We separately show the results for the images that were taken from camera location C2 for the first (outdoor) photo shoot on Table \mbox{\ref{tbl:balcony}}. C2 location was at a balcony of height 230 cm relative to the ground plane where the subjects were standing on. Thus, the camera was pitched down to include the subjects within the field of view. For the other camera locations, the pitch angle was close to zero and people were mainly standing or sitting with their torsos upright. Therefore, the torsos can be expected to be almost parallel to the camera's sensor plane and, thus, produce good distance estimates whenever they are visible. For camera location C2, this may no longer be the case. However, the results show that the relative pair-wise distance estimation errors for C2 locations are only slightly higher than on the average. We can conclude that this level of pitch angle does not cause significant problems. }

 \hl{
A graph showing how the pair-wise distance estimation errors depend on the ground truth distances is given in Fig.~\mbox{\ref{fig:errorsgraph}}. It can be observed from this graph that the pair-wise distance estimations errors are on average slightly lower for higher ground truth distances.}

We also provide additional results by formulating the social distance estimation problem as a binary classification task similar to previous works. We set four different social distance thresholds as safe distances. If the distance between a pair is smaller than the threshold, we consider the distance to be unsafe and safe otherwise. We consider the unsafe case as the positive class. 
The standard evaluation metrics for binary classification problems are Precision, Recall and F1-Score. The formulas for these metrics are $Precision = \frac{True Positives}{True Positives + False Positives}$ , $Recall = \frac{True Positives}{True Positives + False Negatives}$, $F1-score = 2 * (\frac{Precision*Recall}{Precision+Recall})$. F1-score is an overall measure of the binary classification performance and is always within the range of 0-1. An F1-score of 1 indicates perfect classification performance. The F1-score results of our proposed method are given in Table \ref{tbl:f1scores}. \par

\begin{table}[t]
\centering
\begin{tabular}{|c|c|}
\hline
Safe Distance (m) & F1-Score \\ \hline
1                 & 0.39     \\
1.5               & 0.61     \\
2                 & 0.81     \\
3                 & 0.90     \\ \hline
\end{tabular}
\caption{F1-scores of our proposed method for different safe distance thresholds}
\label{tbl:f1scores}
\end{table}

As can be seen in Table \ref{tbl:f1scores}, the choice of safe distance threshold changes the F1-scores drastically. For example, the low performance for 1m threshold follows from many ground-truth distances being just slightly above the threshold. As our methods tends to slightly underestimate the distances as explained in Section~\ref{sec:method}, these cases lead to false positives. This supports our claim that formulating the problem of social distance estimation as a binary classification task is not an optimal way to evaluate the performance of the methods. As the results depend greatly on the threshold value, F1-scores do not reflect the true capacity and accuracy of the distance estimation performance of a method. Our proposed evaluation protocol, which gives the average pair-wise percentual distance estimation error offers greater insight on the method's performance.

\section{Conclusion}
\label{sec:conclusion}
To address the need for more accurate estimation of social distances \hl{from general images to analyze social and cultural impacts of the social distancing regulations introduced due to}  the COVID-19 pandemic, we proposed a new test benchmark for automatic social distance estimation algorithms. The benchmark includes an evaluation protocol for methods producing pair-wise social distances. \hl{The images follow a typical journalistic photographing style instead of a fixed monitoring setup, and they were} taken with \hl{ varying camera} settings. Furthermore, we proposed a robust method that estimates 3D locations of persons in images and then uses these estimated locations to calculate the social distances between the people. \hl{Our method is able to estimate social distances in any single image without the need for knowing the extrinsic parameters or manually calibrating the homography matrix of the image plane to the ground plane, provided that the focal length and sensor size information of the camera are known, which enables our method to be used flexibly on all kinds of images. The proposed} method was able to obtain 92\% person detection rate along with 28.9\% pair-wise distance error on the proposed test benchmark. 

While our method gives satisfactory results for overall shots where the torsos of the people can be detected by OpenPose, the accuracy of the estimations gets weaker for close-up shots where the torsos are generally not visible in the image. \hl{This happens because our method assumes one of the keypoint pairs (eyes, shoulders, torso) to be parallel to the camera's sensor plane, and violations of this assumption lead to distance estimates that are longer than the ground-truth. In typical journalistic photos, where the camera's pitch angle is close to zero and the peoples' torsos are in upright positions, the assumption is typically most accurate for the torso keypoint pair whenever it is visible in the image. Thus, our method could be improved by estimating automatically also the pitch angle and persons' angles with respect to the camera.} Our method also uses average adult human body proportions for the calculations. Therefore, the estimations made for children in the images would be less accurate. \hl{Our method can be improved by taking advantage of other methods that can estimate the gender and ages of the subjects and adaptively changing the assumed body dimensions for each individual subject depending on their gender and age.} It should also be noted that our method requires the focal length and sensor plane size information of the camera. Therefore, our method cannot be applied on photos where these information are lacking. For our method to be applied on pictures where the focal length and sensor plane size are not known, these information would have to be estimated through other methods. 
 \par

In our future research, we will use our benchmark to further enhance the proposed method and then use it in an interdisciplinary study, where we will analyze the impacts of the COVID-19 \hl{regulations on} social interactions. While the COVID-19 makes the social distance analysis very topical, the benchmark and the developed methods are naturally not restricted on COVID-19 related analysis, but they can be beneficial in other image-based \hl{proxemics} studies focusing on different historical, cultural, or journalistic phenomena.


\medskip

\bibliographystyle{IEEEbib}
\bibliography{refs}

\end{document}